\RequirePackage{fix-cm}
\documentclass[smallextended]{svjour3}       
\def\makeheadbox{\relax}  
\smartqed  
\usepackage{placeins}
\usepackage{graphicx}
\usepackage{cite}
\usepackage{amsmath,amssymb,amsfonts}
\usepackage{algorithmicx}
\usepackage{graphicx}
\usepackage{textcomp}
\usepackage{xcolor}
\usepackage{subcaption}
\usepackage{multirow}
\usepackage{comment}
\usepackage[]{todonotes}
\usepackage{longtable,tabu}
\usepackage{layouts}
\usepackage{tabularx}
\newcommand{\shortsection}[1]{\vspace*{1ex}\noindent{\bf #1}}
\usepackage[linesnumbered,ruled]{algorithm2e}
\usepackage[noend]{algpseudocode}
\usepackage{textgreek}
\usepackage[shortcuts]{extdash}
\usepackage[normalem]{ulem}
\usepackage{hyperref}

\usepackage[top=1in,bottom=1in,left=1.25in,right=1.25in]{geometry}

\usepackage{titlesec}
\titlespacing{\section}{0pt}{2em}{1em}
\titlespacing{\subsection}{0pt}{2em}{1em}
\titlespacing{\subsubsection}{0pt}{1em}{1em}
\usepackage{enumitem}
\setlist{nosep}

\newcommand*{\affmark}[1][*]{\textsuperscript{#1}}
\begin{document}

\title{Analysis of Executional and Procedural Errors in Dry-lab Robotic Surgery Experiments}
\author{\mbox{Kay~Hutchinson\affmark[*]\affmark[1]\and
        Zongyu~Li\affmark[*]\affmark[1]\and
        Leigh~A.~Cantrell\affmark[2]\and}
        \mbox{Noah~S.~Schenkman\affmark[3]\and
        Homa~Alemzadeh\affmark[1]}
}

\authorrunning{Hutchinson et al.} 

\institute{ \affmark[*] Co-first-authors: Contributed equally to the paper.\\
            \affmark[1] Department of Electrical and Computer Engineering, \\
            \affmark[2] Department of Obstetrics and Gynecology,\\ 
            \affmark[3] Department of Urology,\\ 
            University of Virginia, Charlottesville, VA 22903 \at
         \email{\affmark[1]\{kch4fk, zl7qw, ha4d\}@virginia.edu,
         \affmark[2,3]\{lac6vz, nss2f\}@hscmail.mcc.virginia.edu}
}

\date{}

\maketitle

\vspace{-5em}

\begin{abstract}  
\hfill \break

\noindent\textbf{Background} \textcolor{black}{Analyzing kinematic and video data can help identify potentially erroneous motions that lead to sub-optimal surgeon performance and safety-critical events in robot-assisted surgery.} 

\noindent\textbf{Methods} We develop a rubric for identifying task and gesture-specific Executional and Procedural errors and evaluate dry-lab demonstrations of Suturing and Needle Passing tasks from the JIGSAWS dataset. We characterize erroneous parts of demonstrations by labeling video data, and use distribution similarity analysis and trajectory averaging on kinematic data to identify parameters that distinguish erroneous gestures.

\noindent\textbf{Results} Executional error frequency varies by task and gesture, and correlates with skill level. Some predominant error modes in each gesture are distinguishable by analyzing error-specific kinematic parameters. Procedural errors could lead to lower performance scores and increased demonstration times but also depend on surgical style.

\noindent\textbf{Conclusions} \textcolor{black}{This study provides insights into context-dependent errors that can be used to design automated error detection mechanisms and improve training and skill assessment.} 

\end{abstract}

\section{Introduction} \label{intro}
\noindent With advances in sensing and computing technology, artificial intelligence, and data science, the next generation of Robot-Assisted Surgery (RAS) systems is envisioned to benefit from new capabilities for context-specific monitoring~\cite{yasar_dsn2020} and virtual coaching during simulation training as well as decision support and cognitive assistance during actual surgery to improve safety, efficiency, and quality of care\cite{taylor2016medical}. State-of-the-art RAS systems and simulators are designed with data logging mechanisms to collect system logs, kinematics, and video data from surgical procedures. The recorded data has been mostly used for offline surgical skill evaluation \cite{reiley2009decomposition, rosen2006generalized, fard2018automated}, with the aim of improving surgeons' performance and making evaluations objective and scalable.

Current methods for objective assessment of robotic technical skills can be classified into two general categories: manual assessment and automated assessment. 
Manual skill evaluation is usually performed globally, assessing performance over an entire demonstration using frameworks such as OSATS (Objective Structured Assessment of Technical Skills) \cite{martin1997objective}, GOALS (Global Operative Assessment of Laparoscopic Skills) \cite{vassiliou2005global}, GEARS \cite{sanchez2016robotic}, and R-OSATS \cite{Tarr2014ROSATS}. 
However, manual assessment methods are subjective, cognitively demanding, and prone to errors \cite{fard2018automated}. 
In response, automated assessment methods utilizing kinematic, video, and system event data \cite{Azizian2020fusion} are being developed to provide objective and quantitative metrics \cite{fard2018automated} and \cite{Chen2019ObjectiveAO}, and explainable feedback \cite{fawaz2019accurate}. 
Automated methods also allow the subdivision of demonstrations into subtasks or gestures, and to base performance assessment and technical skill evaluation on the quality and/or sequence of these components as proposed in \cite{ahmidi2017dataset}, \cite{Tao_Hager_2012}, and \cite{rosen2006generalized}. Further, some gestures are more indicative of skill level than others \cite{Varadarajan_Hager_2009}.

The metrics used for surgical skill assessment can be classified into three broad categories of: i) efficiency (e.g., path length, completion time), ii) safety (e.g., instrument collisions \cite{poursartip2018analysis}, instruments out of view, excessive force, needle drops, tissue damage \cite{chowriappa2013development}), and iii) task/procedure specific metrics (e.g., task outcome metrics, camera movement, energy activation \cite{hung2018development}).

While most previous works focused on skill evaluation for distinguishing between expertise levels, less attention has been paid to identifying specific erroneous surgical motions that contribute to sub-optimal performance and potential safety-critical events. 
\textcolor{black}{The closest related works are \cite{moorthy2004bimodal} and \cite{guni2018development} which proposed objective gesture-based checklists for laparoscopic and robot-assisted suturing.}  
Others have proposed general and custom rubrics for evaluation of human errors \cite{joice1998errors} and technical errors \cite{bonrath2013defining} in laparoscopic surgery. Related works on errors in RAS mainly focused on analyzing adverse events and system malfunctions as reported by the surgical teams and institutions\cite{alemzadeh2016adverse}. \textcolor{black}{Augmenting RAS systems and simulators with mechanisms for monitoring the progress of surgical tasks and providing early and context-specific feedback to surgeons on potentially sub-optimal or unsafe motions could help improve performance scores in training and prevent safety-critical events in actual surgery \cite{yasar_dsn2020}.}

\textcolor{black}{In this study, we take a step towards developing automated safety monitoring mechanisms for RAS by defining a rubric for identifying task and gesture-specific errors based on video and kinematic data. We use this rubric to analyze}
recorded dry-lab demonstrations of two common tasks (Suturing and Needle Passing) performed on the da Vinci Surgical System (dVSS).
We focus on identifying which parts of a trajectory (spanning one or more gestures) are potentially erroneous (sub-optimal) versus error-free (optimal). We then characterize the erroneous trajectories by identifying the most common types of errors for each task and gesture, and the kinematic parameters and surgeon-specific signatures that distinguish between optimal and sub-optimal performance. \textcolor{black}{The results from this study can aid in designing more efficient training modules, curricula, and simulation tools that reinforce optimal performance by providing detailed, quantitative, and context-specific feedback to surgeons.
In summary, we propose a novel framework for objective evaluation of RAS procedures with the following key contributions\footnote{\textcolor{black}{Our labels, code, and data are available at \url{https://github.com/UVA-DSA/ExecProc_Error_Analysis}.}}:
\begin{itemize}
    \item A task and gesture-specific rubric for identification of executional and procedural errors using data collected from real or simulated surgical demonstrations.
    \item A set of executional error labels based on manual annotation of video data to augment the Suturing and Needle Passing tasks of the JIGSAWS dataset \cite{gao2014jhu}. 
    \item Quantitative analysis methods to characterize gesture-specific Executional and task-specific Procedural errors using pre-collected kinematic data and gesture labels.
    \item Insights on the types, frequencies, and durations of Executional and Procedural errors across tasks and gestures and their correlations with skill levels which can provide a basis for the design of automated error detection mechanisms. 
\end{itemize}}


\section{Methods}
\label{sec:methods}
\begin{figure}[t!]
    \centering
    \vspace{0.5em}
    \includegraphics[trim = 0.2in 2in 3.9in 0.2in, width=\textwidth]{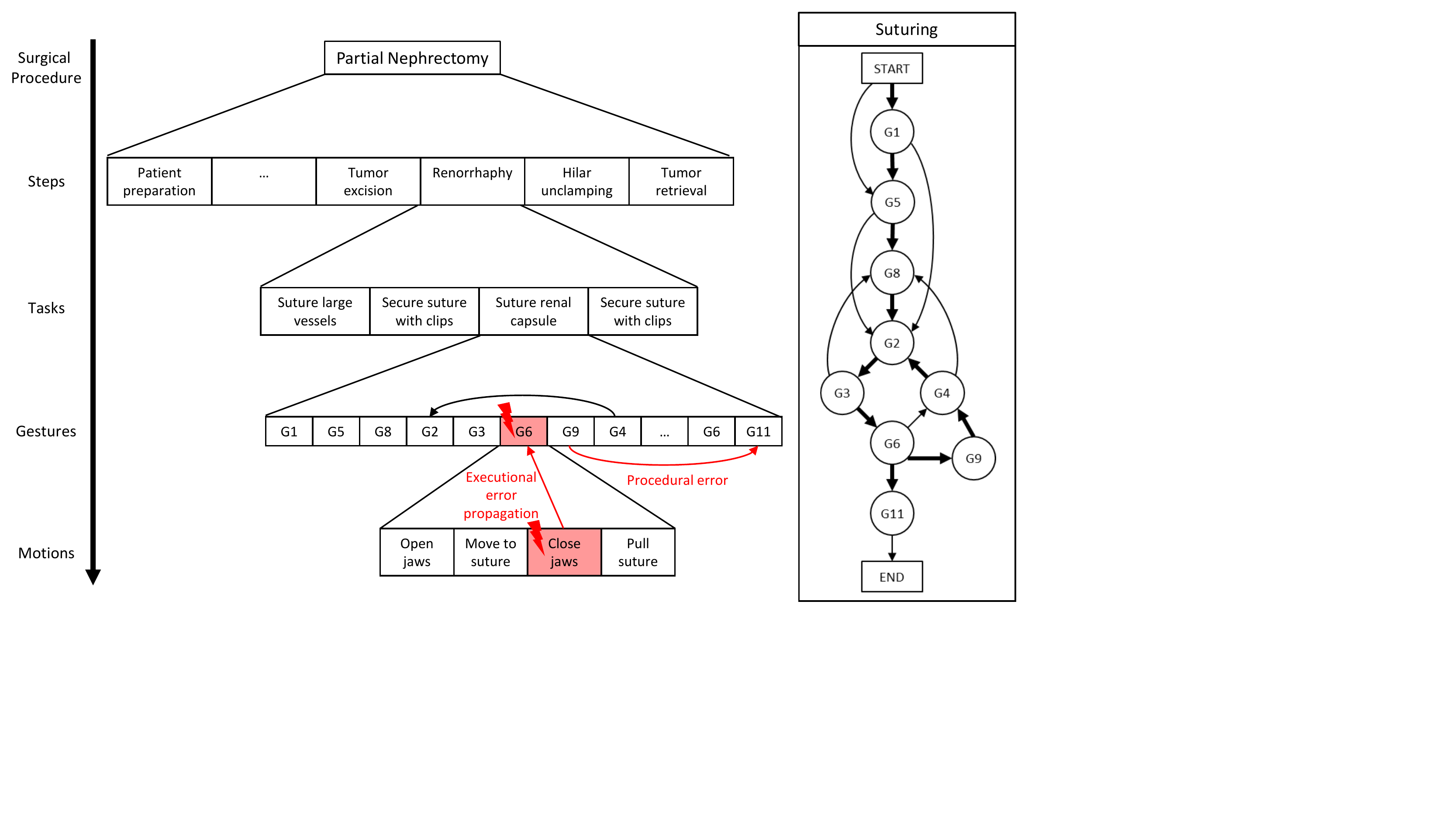}
    \caption{Surgical hierarchy (adopted from \cite{neumuth2011modeling}) for an example urological procedure of partial nephrectomy (based on \cite{sukumar2011robotic} and \cite{kaouk2012robot}) along with example gesture-specific Executional and Procedural Errors}
    \label{fig:hierarchy}
\end{figure}

\noindent Sources of errors in RAS are diverse and domain-specific, including faults in the robotic system software and hardware, or human errors \cite{alemzadeh2016adverse}. In this study, we focus on errors in the execution of procedures that can be observed in video recordings and detected in kinematic data. Surgical procedures follow the hierarchy of levels defined in \cite{neumuth2011modeling} which provides context \cite{yasar_dsn2020} for actions during the operation, as shown in Figure \ref{fig:hierarchy}. A surgical \textbf{operation} can involve multiple \textbf{procedures} which are divided into \textbf{steps}. Each \textbf{step} is subdivided into \textbf{tasks} comprised of \textbf{gestures} (also called sub-tasks or surgemes) which are made of \textbf{motions} such as moving an instrument or closing the graspers. Errors can occur at any level of this hierarchy and can propagate and cause errors at other levels. We specifically focus on studying the quality of the task demonstrations at the \textit{gesture} level to answer the following research questions:

\shortsection{RQ1:} Which tasks and gestures are most prone to errors? 

\shortsection{RQ2:} Are there common error modes or patterns across gestures and tasks? 

\shortsection{RQ3:} Are erroneous gestures distinguishable from normal gestures?  

\shortsection{RQ4:} What kinematic parameters can be used to distinguish between normal and erroneous gestures?

\shortsection{RQ5:} Do errors impact the duration of the trajectory? 

\shortsection{RQ6:} Are there any correlations between errors and surgical skill levels?

\subsection{Rubric for Objective Assessment of Errors in Robotic Surgery}

\textcolor{black}{\noindent Our goal is to define a rubric for identification of errors based on video and/or kinematic data, which can also be used for automated error detection using quantitative measures such as instrument position, amount of force, traveling distance, and system events.} We adopt a previous categorization of human errors in laparoscopic surgery from \cite{joice1998errors} and define two types of errors in our rubric: \textbf{Procedural errors} and \textbf{Executional errors}. Procedural errors involve ``the omission or re-arrangement of correctly undertaken steps within the procedure," while Executional errors are ``the failure of a specific motor task within the procedure." \textbf{Technical errors} are the ``failure of a planned action to achieve a goal", including inadequate (too much/too little) use of force or distance, inadequate visualization and wrong orientation of instruments or dissection plane \cite{Bonrath2013errorRating}, and are considered a subtype of Executional errors that can be quantified with thresholds. 

\begin{table}[t!]
\resizebox{\textwidth}{!}{%
\begin{tabular}{|c|l|l|c|c|c|c|}
\hline
\multicolumn{2}{|l|}{\multirow{2}{*}{\textbf{Gesture Description}}} &
  \multirow{2}{*}{\textbf{Error Mode}} &  \multicolumn{2}{c|}{\textbf{Suturing}} &  \multicolumn{2}{c|}{\textbf{Needle Passing}} \\ \cline{4-7} 
\multicolumn{2}{|l|}{} &   &  \textbf{\begin{tabular}[c]{@{}c@{}}Total\\ No. \\ Errors\end{tabular}} &  \textbf{\begin{tabular}[c]{@{}c@{}}Erroneous\\ Gestures\\ (\%)\end{tabular}} &  \textbf{\begin{tabular}[c]{@{}c@{}}Total\\ No.\\ Errors\end{tabular}} &  \textbf{\begin{tabular}[c]{@{}c@{}}Erroneous\\ Gestures\\ (\%)\end{tabular}} \\ \hline
\multirow{3}{*}{\textbf{G1}} &  \multirow{3}{*}{\begin{tabular}[c]{@{}l}Reaching for needle \\with right hand\end{tabular}} &  Multiple attempts &  7 & \multirow{3}{*}{\begin{tabular}[c]{@{}c@{}}8/29\\ (28\%)\end{tabular}} &  N/A &  \multirow{3}{*}{\begin{tabular}[c]{@{}c@{}}11/30\\ (37\%)\end{tabular}} \\ \cline{3-4} \cline{6-6} &   &  Needle drop &  0 &   &  2 &   \\ \cline{3-4} \cline{6-6} &   &  Out of view &  1 &   &  10 &   \\ \hline
\multirow{3}{*}{\textbf{G2}} &  \multirow{3}{*}{Positioning needle} &  Multiple attempts &  21 &  \multirow{3}{*}{\begin{tabular}[c]{@{}c@{}}22/166\\ (13\%)\end{tabular}} &  51 &  \multirow{3}{*}{\begin{tabular}[c]{@{}c@{}}55/117\\ (47\%)\end{tabular}} \\ \cline{3-4} \cline{6-6} &   &  Needle drop &  0 &   &  0 &   \\ \cline{3-4} \cline{6-6} &   &  Out of view &  1 &   &  6 &   \\ \hline
\multirow{3}{*}{\textbf{G3}} &  \multirow{3}{*}{\begin{tabular}[c]{@{}l}Pushing needle \\through tissue\end{tabular}} &
  \begin{tabular}[c]{@{}l@{}}Not moving along the curve/\\ Multiple attempts\end{tabular} &  80 & \multirow{3}{*}{\begin{tabular}[c]{@{}c@{}}82/164\\ (51\%)\end{tabular}} &
  17 &  \multirow{3}{*}{\begin{tabular}[c]{@{}c@{}}17/111\\ (15\%)\end{tabular}} \\ \cline{3-4} \cline{6-6} &   &  Needle drop &  0 &   &  0 &   \\ \cline{3-4} \cline{6-6} &   &  Out of view &  2 &   &  0 &   \\ \hline
\multirow{4}{*}{\textbf{G4}} &  \multirow{4}{*}{\begin{tabular}[c]{@{}l}Transferring needle \\from left to right\end{tabular}} &  Multiple attempts &  19 & \multirow{4}{*}{\begin{tabular}[c]{@{}c@{}}71/119\\ (60\%)\end{tabular}} &
  15 &  \multirow{4}{*}{\begin{tabular}[c]{@{}c@{}}23/83\\ (28\%)\end{tabular}} \\ \cline{3-4} \cline{6-6} &   &  Needle orientation &  53 &   &  9 &   \\ \cline{3-4} \cline{6-6}&   &  Needle drop &  0 &   &  0 &   \\ \cline{3-4} \cline{6-6} &   &  Out of view &  14 &   &  3 &   \\ \hline
\multirow{2}{*}{\textbf{G5}} &  \multirow{2}{*}{\begin{tabular}[c]{@{}l}Moving to center \\with needle in grip\end{tabular}} &  Needle drop &  1 &  \multirow{2}{*}{\begin{tabular}[c]{@{}c@{}}2/37\\ (5\%)\end{tabular}} &  0 &  \multirow{2}{*}{\begin{tabular}[c]{@{}c@{}}3/31\\ (10\%)\end{tabular}} \\ \cline{3-4} \cline{6-6} &   &  Out of view & 1 &   &  3 &   \\ \hline
\multirow{3}{*}{\textbf{\begin{tabular}[c]{@{}c@{}}G6\end{tabular}}} &  \multirow{3}{*}{\begin{tabular}[c]{@{}l}Pulling suture with \\left hand\end{tabular}} &  Multiple attempts & 8 &  \multirow{3}{*}{\begin{tabular}[c]{@{}c@{}}121/163\\ (74\%)\end{tabular}} & 14 &  \multirow{3}{*}{\begin{tabular}[c]{@{}c@{}}46/112\\ (41\%)\end{tabular}} \\ \cline{3-4} \cline{6-6} &   &  Needle drop &  2 &   &  0 &   \\ \cline{3-4} \cline{6-6} &   &  Out of view &  120 &   &  37 &   \\ \hline
\multirow{4}{*}{\textbf{G8}} &  \multirow{4}{*}{Orienting needle} &  Multiple attempts &  18 &  \multirow{4}{*}{\begin{tabular}[c]{@{}c@{}}28/48\\(58\%)\end{tabular}} &  1 &  \multirow{4}{*}{\begin{tabular}[c]{@{}c@{}}3/28\\ (11\%)\end{tabular}} \\ \cline{3-4} \cline{6-6} &   &  Needle orientation & 22 &   &  1 &   \\ \cline{3-4} \cline{6-6} &   &  Needle drop &  0 &   &  0 &   \\ \cline{3-4} \cline{6-6} &   &  Out of view &  4 &   &  2 &   \\ \hline
\multirow{3}{*}{\textbf{G9}} &  \multirow{3}{*}{\begin{tabular}[c]{@{}l}Using right hand to \\help tighten suture\end{tabular}} &  Multiple attempts & 3   & \multirow{3}{*}{\begin{tabular}[c]{@{}c@{}}11/24\\ (46\%)\end{tabular}} & 1 &  \multirow{3}{*}{\begin{tabular}[c]{@{}c@{}}1/1\\ (100\%)\end{tabular}} \\ \cline{3-4} \cline{6-6} &   &  Needle drop &  0 &   &  1 &   \\ \cline{3-4} \cline{6-6} &   &  Out of view &  11 &   &  0 &   \\ \hline

\multirow{4}{*}{\textbf{\begin{tabular}[c]{@{}c@{}}All \\gestures\end{tabular}}} &  \multirow{4}{*}{\begin{tabular}[c]{@{}l}Total number of errors \\across all gestures\end{tabular}} &  Multiple attempts & 156 & \multirow{4}{*}{\begin{tabular}[c]{@{}c@{}}345/750\\(46\%)\end{tabular}} &  99 &  \multirow{4}{*}{\begin{tabular}[c]{@{}c@{}}159/513\\(31\%)\end{tabular}} \\ \cline{3-4} \cline{6-6} &   &  Needle drop &  3  &   &  3  &   \\\cline{3-4} \cline{6-6} &   &  Needle orientation &  75  &   &  10  &   \\ \cline{3-4} \cline{6-6} &   &  Out of view &  154  &   &  61  &   \\ \hline
\end{tabular}%
}
\caption{Gesture-specific Executional errors for Suturing and Needle Passing in the JIGSAWS dataset. \textcolor{black}{Example videos for each error mode can be found at \url{https://github.com/UVA-DSA/ExecProc_Error_Analysis}}.}
\label{tab:all_errors}
\end{table}

In order to generalize these definitions to different procedures and tasks, we define Executional and Procedural errors at the gesture level. More specifically, we define a set of Executional error modes for each gesture as listed in the rubric in Table \ref{tab:all_errors}. Some errors are gesture-specific such as ``Needle orientation" which is only defined for G4 and G8 as those gestures specifically manipulate the needle in preparation for positioning the needle (G2) and throwing the next suture (G3), as shown in the grammar graph of Figure \ref{fig:hierarchy} (adopted from \cite{ahmidi2017dataset}). The standard acceptable practice for those gestures is to hold the needle in the grasper 1/2 to 2/3 of the way from the tip of the needle and with the needle perpendicular to the jaws of the grasper \cite{moorthy2004bimodal}. 
Other gestures that do not purposely alter the orientation of the needle in the grasper cannot have this error mode. For G3, the definition of a ``Multiple attempts" error also includes ``Not moving along the curve" of the needle (from \cite{moorthy2004bimodal}) since these two errors are very difficult to distinguish and often happen simultaneously. Other error modes, including ``Multiple attempts", ``Needle drop", and ``Out of view", could occur at any time during a task and are considered for every gesture. 

We define Procedural errors as any deviation in the sequence of gestures performed in a demonstration from the standard accepted gesture sequences defined for that task and shown in the grammar graphs in Figures \ref{fig:hierarchy} and \ref{fig:analysisflowchart}. \cite{joice1998errors} defined several sub-categories for Procedural errors, including adding an unexpected step, skipping a step, out of order transition, and repetition of steps. These subcategories are included in our analysis of Procedural errors as discussed in Section \ref{sec:procedural_errors}.

\subsection{JIGSAWS Dataset}
\label{background}
\noindent The JHU-ISI Gesture and Skill Assessment Working Set (JIGSAWS)~\cite{gao2014jhu} is a publicly available dataset, \textcolor{black}{collected using the Research API for the da Vinci Surgical System (dVSS)} from eight surgeons of varying skill levels performing three dry-lab surgical tasks: Suturing, Knot Tying, and Needle Passing. These tasks are among the standard modules in most surgical skills training curricula. 

The JIGSAWS dataset includes kinematic and video data from up to 39 demonstrations (or trials) per task along with manually annotated gesture \textbf{transcripts} (indicating the sequence of gestures, with the beginning and end of each gesture and its type) and surgical skill levels for each demonstration. The vocabulary of surgical gestures used for labeling is shown in Table \ref{tab:all_errors}. Surgical skills were characterized using both self-proclaimed expertise levels and Global Rating Scale (GRS) score for each demonstration. Self-proclaimed (SP) expertise levels were based on the number of hours of robotic surgical experience, divided into: \textbf{SP-Expert} ($>$100 hrs), \textbf{SP-Intermediate} (10-100 hrs), and \textbf{SP-Novice} ($<$10 hrs). GRS scores were given using a modified Objective Structured Assessments of Technical Skills (OSATS) approach based on 6 elements (on a rating-scale of 1-5 per element): Respect for tissue, Suture/needle handling, Time and motion, Flow of operation, Overall performance and Quality of final product \cite{gao2014jhu}. 
We also classified the demonstrations into three groups based on the GRS scores: \textbf{GRS-Novice} (0$\le$GRS $\le$9), \textbf{GRS-Intermediate} (10$\le$GRS $\le$19), and \textbf{GRS-Expert} (20$\le$GRS $\le$30).

\begin{figure}[t!]
    \centering
    \vspace{0em}
    \includegraphics[trim = 0in 1.8in 7in 0in, width=\textwidth]{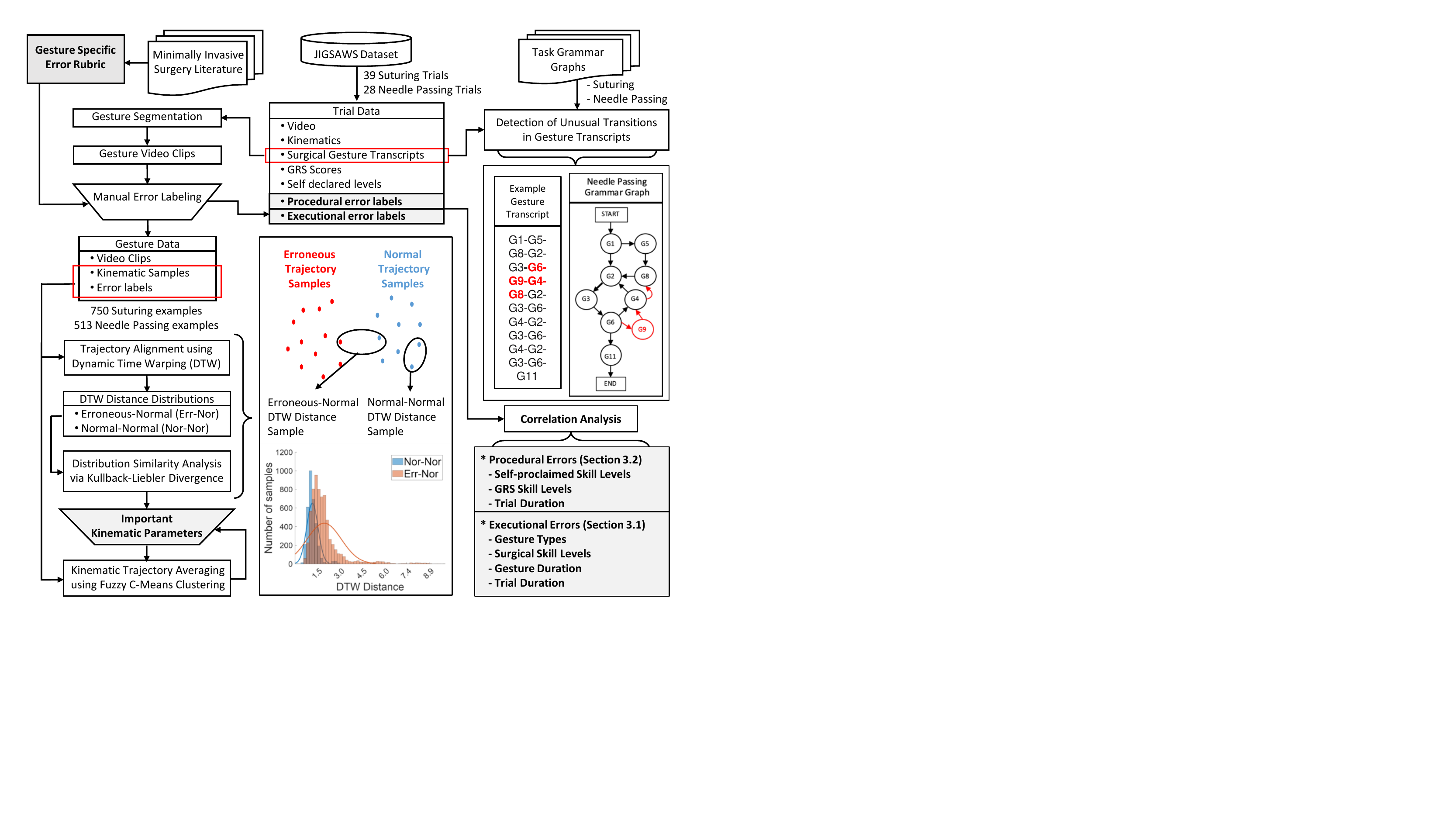}
    \caption{Overall Methodology for Analysis of Executional and Procedural Errors}
    \vspace{-2em}
    \label{fig:analysisflowchart}
\end{figure}

Figure \ref{fig:analysisflowchart} shows our overall pipeline for the analysis of Executional and Procedural errors in the JIGSAWS dataset. Due to the limited number of demonstrations for the Knot Tying task in the dataset, our analysis only focused on Suturing and Needle Passing.

\subsection{Executional Error Analysis}
\noindent Kinematic and video data for each trial were first segmented into gestures based on the gesture transcript annotations. The video clip for each gesture was then reviewed and labeled by two to three independent annotators \textcolor{black}{(with experience in robotic surgery and/or suturing)} as normal or erroneous for each error mode. Final labels for each error mode were obtained by taking the consensus among annotators. 
A gesture example that exhibited one or more errors was marked as erroneous, otherwise, it was labeled as normal. We then proceeded with the analysis of the patient-side manipulator (PSM) kinematic data corresponding to each gesture for all the normal and erroneous demonstrations of each task. 

\subsubsection{Dynamic Time Warping}
\noindent We used Dynamic Time Warping (DTW) to measure the similarity between normal and erroneous trajectories for each gesture. DTW is an effective method for aligning two temporal sequences, independent of the non-linear variations in time, by minimizing the Euclidean distance between the two signals. In our analysis, we performed independent DTW on each variable before summing the returned distances for each parameter listed in Table \ref{tab:JIGSAWS_params}. We found no significant difference between this method and dependent DTW where all variables in each parameter group were warped together yielding a single distance instead of a sum of distances (similar observations were made in \cite{shokoohi2015non}). 
DTW was performed on every combination of two example trajectories for each gesture. From this, we obtained comparisons of normal examples to other normal examples (``Nor-Nor") and comparisons of erroneous examples to normal examples (``Err-Nor"). The DTW distance samples represented a distribution of distances for the ``Nor-Nor" and ``Err-Nor" subsets as shown in the histogram of Figure \ref{fig:analysisflowchart}.
 This resulted in two sets of distance samples for each parameter, each representing a DTW distribution for a comparison subset. 

\begin{table}[]
\centering
\resizebox{\textwidth}{!}{%
\begin{tabular}{|l|l|l|l|}
\hline
\textbf{Index} & \textbf{Description of variables} & \textbf{Parameter name} \\ \hline
39-41 & Right PSM1 tool tip position (xyz) & R Pos \\ \hline
42-50 & Right PSM1 tool tip rotation matrix (R) & R Rot Mat \\ \hline
51-53 & Right PSM1 tool tip linear velocity (x' y' z') & R Lin Vel \\ \hline
54-56 & Right PSM1 tool tip rotational velocity (\textalpha' \textbeta' \textgamma') & R Rot Vel \\ \hline
57 & Right PSM1 gripper angle (\textTheta) & R Grip Ang \\ \hline
58-60 & Left PSM2 tool tip position (xyz) & L Pos \\ \hline
61-69 & Left PSM2 tool tip rotation matrix (R) & L Rot Mat \\ \hline
70-72 & Left PSM2 tool tip linear velocity (x' y' z') & L Lin Vel \\ \hline
73-75 & Left PSM2 tool tip rotational velocity (\textalpha' \textbeta' \textgamma') & L Rot Vel \\ \hline
76 & Left PSM2 gripper angle (\textTheta) & L Grip Ang \\ \hline
\end{tabular}%
}
\caption{Kinematic variables in the JIGSAWS dataset (adopted from \cite{gao2014jhu})}
\label{tab:JIGSAWS_params}

\end{table}

\subsubsection{Kullback-Liebler Divergence}
\noindent Kullback-Liebler (KL) Divergence, also called relative entropy, is a non-symmetric measure of the difference between two probability distributions. 
The KL Divergence between two identical distributions is zero. As shown in Equation \ref{equation:KL Divergence}, KL Divergence was used to compare the ``Err-Nor" and ``Nor-Nor" DTW distance distributions for each gesture to determine which parameters had a significant difference between the two distributions. 

\begin{equation}
    D_{KL}(DTW_{Err-Nor} || DTW_{Nor-Nor}) = -\Sigma DTW_{Err-Nor} log\left(\frac{DTW_{Nor-Nor}}{DTW_{Err-Nor}}\right)
    \label{equation:KL Divergence}
\end{equation}

\subsubsection{Trajectory Averaging}
\noindent We examined the kinematic data for important parameters to verify differences between normal and erroneous gestures using a method based on \cite{hu2006system}.
Each signal was time-normalized by downsampling the signal by 3 (keeping only every third sample) and then linearly interpolated to stretch it to the average duration of the normal or erroneous gesture examples of that task 
(supported by our analysis of gesture durations in Section \ref{sec:exec_errors_durations}). Then, fuzzy c-means clustering was performed on each variable and its normalized time index to obtain the average normal and erroneous trajectories (represented by 15 cluster centers), shown with blue (normal) and red (erroneous) dots in Figure \ref{fig:S_G1_57}. 

\subsection{Procedural Error Analysis}
\label{sec:procedural_errors}
\noindent Previous works proposed modeling the standard acceptable gesture sequences for a task using a grammar graph that shows the relationship, order, and flow of gestures \cite{Varadarajan_Hager_2009},
\cite{ahmidi2017dataset}.
\cite{Tao_Hager_2012}. 
The grammar graph of a task is a digraph with each vertex representing the set of 
gestures for the task and each edge representing a common transition between two gestures. We adopted the grammar graphs for Suturing and Needle Passing from \cite{ahmidi2017dataset} and included an additional directed link from G1 to G2 in Suturing (see Figures \ref{fig:hierarchy} and \ref{fig:analysisflowchart}). 

\par We acquired the gesture sequences performed for Suturing and Needle Passing from the JIGSAWS transcripts. 
Then we developed a method for checking if each gesture sequence follows the standard acceptable sequence of gestures in the grammar graph. As shown in Algorithm \ref{alg:procedure_error_algorithm_K}, for each gesture we check if it is in the grammar graph for the task and if it is a valid successor of the previous gesture, otherwise it is marked as a procedural error. 
Each transcript can have multiple, possibly sequential, procedural errors. This algorithm, combined with a gesture segmentation algorithm, can be used for automated detection of procedural errors in real-time.

Deviations from the grammar graph might also happen because of variations in surgical style and expertise, as discussed in Section \ref{sec:procedural_errors_results}.

\begin{algorithm}[t!]
 \footnotesize
 \DontPrintSemicolon
    \textbf{Input:} \newline
    - A grammar graph $G(V,E)$ for a surgical task which is a digraph with each vertex in $V$ representing the entry point START or one of the common gesture types $G_{i}$ in the task, and each $edge(G_{i},G_{j}) \in E$ representing a common transition between gestures $G_{i}$ and $G_{j}$.\newline
    - A set of $m$ task transcripts $T=\{T_{1},T_{2},...,T_{m}\}$, where $T_{k} \in T$ is an ordered sequence of $n$ gestures $T_{k}=[G_{1},G_{2},...,G_{n}]$
    \newline
    \textbf{Output:} \newline
    - A list of erroneous gesture transitions $error\_seq$ for each transcript
    \newline
    \For {$T_{k} \in T$} 
    {
         $error\_seq = \emptyset$ \newline
         $val \gets G.successors(START)$ \newline
         \For{$G_{i} \in T_k$}{
            \uIf {$G_{i} \in V$}
            {
                \uIf{$G_{i} \notin val$}{ 
                    $error\_seq$.append([$G_{i-1},G_{i}$])}
                $val \gets G.successors(G_{i})$
            } 
            \lElse{ 
                $error\_seq$.append([$G_{i}$]) \newline
                $val \gets [G_{i+1}]$ 
            }
        }
    }
    \caption{Procedural Error Detection Algorithm}
    \label{alg:procedure_error_algorithm_K}

\end{algorithm}
\setlength{\textfloatsep}{2em}

\section{Results}
\noindent This section presents a summary of results and observations from our analysis of the JIGSAWS dataset. 

\textbf{\subsection{Executional Errors}}

\noindent Table \ref{tab:all_errors} lists the number of examples of each error mode as well as the total number of erroneous examples for each gesture. Note that a gesture example could exhibit multiple error modes, so the sum of the total number of errors does not necessarily equal the number of erroneous gestures.\\ 

\subsubsection{Distribution of Executional Errors Among Gestures}
\noindent Figure \ref{fig:errdist} shows the distribution of errors of each type for each gesture from Table \ref{tab:all_errors}. If a gesture example had more than one error label, it was counted under the ``Multiple errors" category.  
We made the following observations:

\begin{figure}
    \centering
    \vspace{-1em}
    \includegraphics[trim = 0.1in 2.5in 0.1in 2.5in, width=\textwidth]{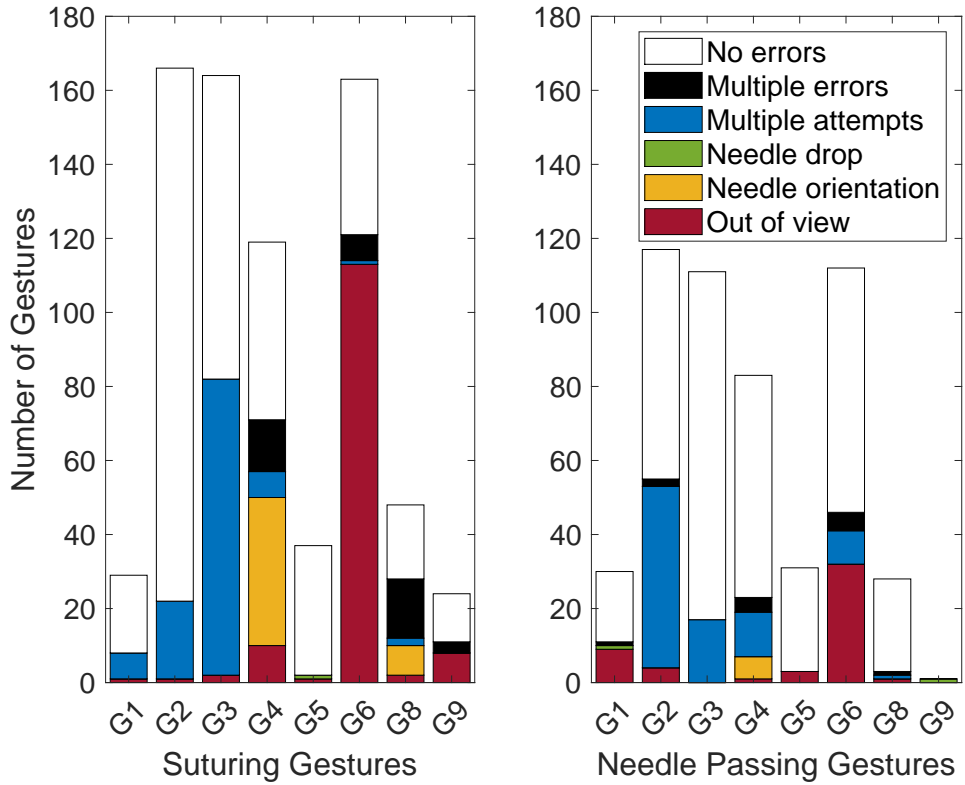}
    \caption{Distribution of Errors for Each Gesture}
    \label{fig:errdist}
    
\end{figure}

\begin{itemize}
    \item G5 for both tasks and G1, G8, and G9 for Needle Passing did not have enough examples of executional errors, so further analysis was not performed on these gestures. G8 from Needle Passing and G5 from both tasks had the lowest percentage of errors because they may be less challenging than other gestures.
    
    \item G2 and G3 have the most ``Multiple attempts" errors in both tasks because they require a high level of accuracy in positioning and driving the needle though the tissue, respectively. G2 has more errors in Needle Passing because the eye of the ring is a smaller target than the dot on the fabric. G3 has more errors in Suturing because surgeons often tried multiple times to align the tip of the needle with the exit point while the needle was not visible beneath the fabric. Comparatively in Needle Passing, the needle only had to pass through one point and was always visible.
    
    \item G4 and G6 from both tasks, and G8 from Suturing have the most gestures with multiple errors. G4 and G8 both involve manipulating the needle between the graspers and the predominant error modes were ``Needle orientation" and ``Multiple attempts" likely due to issues with hand coordination. For G6, the main error modes were ``Out of view" and ``Multiple attempts" due to multiple attempts at grasping the needle and pulling it through the ring or tissue and then moving off-camera to pull the suture through. 
    
    \item G6 has a large number of ``Out of view" errors especially in Suturing possibly because surgeons could not move the camera for the trials in the JIGSAWS data set. However, a different technique to pull the suture could have been used such as hand-over-hand or the pulley method that would have kept the tools within view.
    
\end{itemize}

\subsubsection{Kinematic Parameters for Distinguishing Errors in Each Gesture}
\label{section:kin_params}
\noindent We performed a comparative analysis of KL Divergence values for parameters in each gesture and identified the kinematic parameters that are associated with error occurrence as listed in Table \ref{tab:kin_params}. The following are key observations from this analysis:

\begin{table}[]
\centering
\resizebox{0.6\textwidth}{!}{%
\begin{tabular}{|l|l|l|}
\hline \textbf{Task}  & \textbf{Gesture} & \textbf{Parameters}  \\ \hline \multirow{5}{*}{Suturing}       & G1      & \begin{tabular}[c]{@{}l@{}}Right Gripper Angle\\ Right Linear Velocity\\ Right Position\end{tabular}  \\ \cline{2-3} 
& G3 & \begin{tabular}[c]{@{}l@{}}Right Linear Velocity\\ Right Rotational Velocity\\ Right Gripper Angle\end{tabular}  \\ \cline{2-3} & G6 & Left Position \\ \cline{2-3} 
& G8      & \begin{tabular}[c]{@{}l@{}}Right Position\\ Left Gripper Angle\\ Left Linear Velocity\\ Right Gripper Angle\end{tabular} \\ \cline{2-3} 
& G9      & Left Gripper Angle     \\ \hline
\multirow{2}{*}{Needle Passing} & G2      & \begin{tabular}[c]{@{}l@{}}Left Rotational Velocity\\ Left Linear Velocity\end{tabular}   \\ \cline{2-3} 
 & G3      & \begin{tabular}[c]{@{}l@{}}Left Rotational Velocity\\ Right Rotation Matrix\\ Right Gripper Angle\end{tabular} \\ \hline
\end{tabular}%
}
\caption{Kinematic Parameters with the Greatest KL Divergence Distinguishing Errors in Different Gestures}
\label{tab:kin_params}

\end{table}

\begin{itemize}
    \item For G1 in Suturing, the predominant error mode was ``Multiple attempts" at picking up the needle. Figures \ref{fig:S_G1_57} and \ref{fig:S_G1_Rpos} show that erroneous gestures exhibited a second opening and closing of the grasper and a large difference in Y Position trajectories. This explains the large KL Divergences for those right hand parameters in Figure \ref{fig:S_G1_KLD}.

\begin{figure}
    \centering
    \vspace{-6em}
    \begin{subfigure}[b]{0.49\textwidth}
        \centering
        \includegraphics[trim = 0.3in 1.5in 0.3in 1.5in, width=\textwidth]{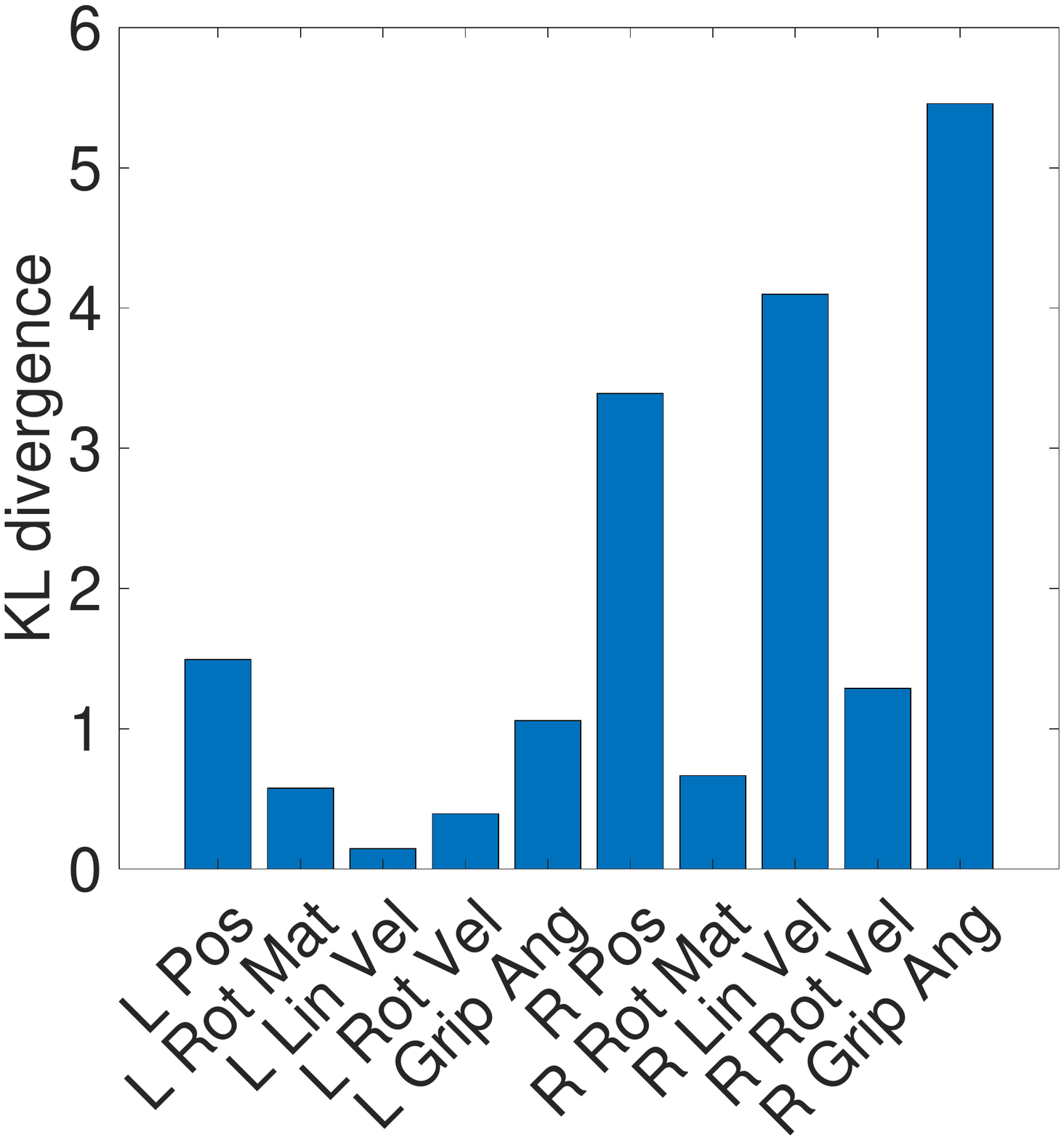}
        \caption{}
        \label{fig:S_G1_KLD}
    \end{subfigure}
    \hfill
    \begin{subfigure}[b]{0.49\textwidth}
        \centering
        \includegraphics[trim = 0.3in 1.05in 0.3in 0in, width=0.98\textwidth]{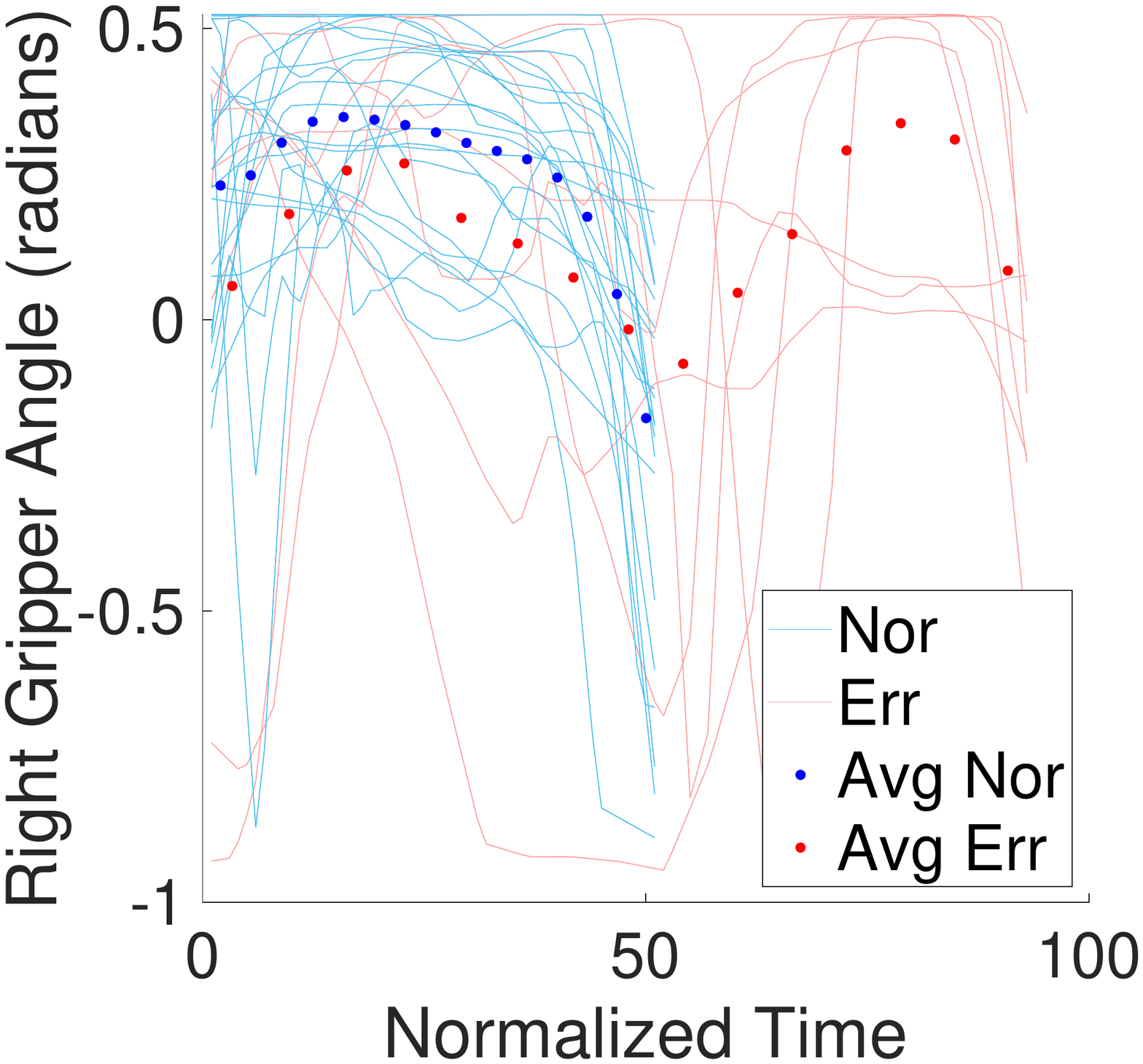}
        \caption{}
        \label{fig:S_G1_57}
    \end{subfigure}
    \vspace{-1em}
    \caption{G1 in Suturing: (a) KL Divergence of Kinematic Parameters, (b) Right Gripper Angle Trajectories for Normal and Erroneous Gestures}
    \label{fig:S_G1}
\end{figure}

\begin{figure}
    \centering
    \includegraphics[trim = 0.1in 2.5in 0.1in 2.5in, width=0.32\textwidth]{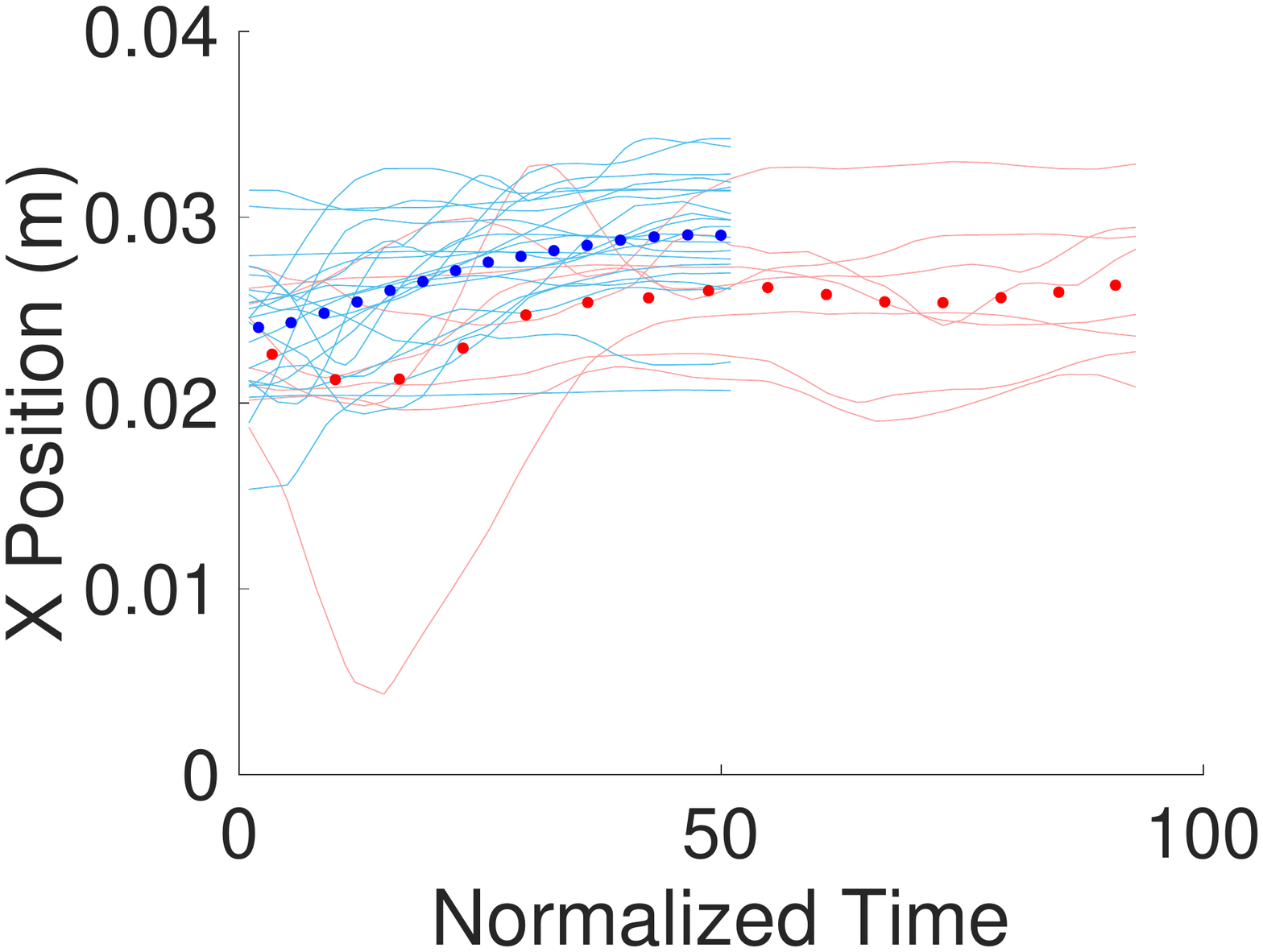}
    \includegraphics[trim = 0.1in 2.5in 0.1in 2.5in, width=0.32\textwidth]{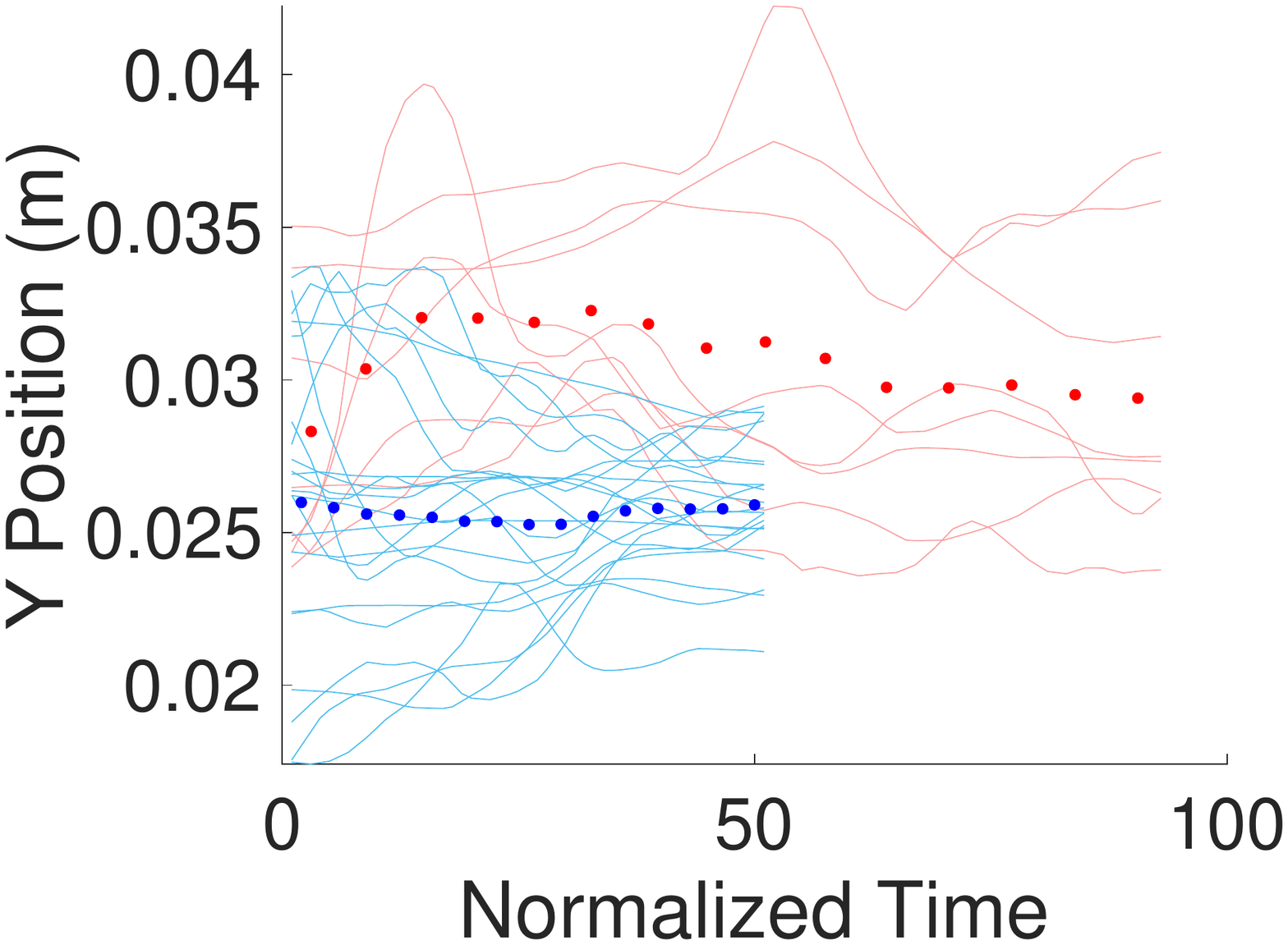}
    \includegraphics[trim = 0.1in 2.5in 0.1in 2.5in, width=0.32\textwidth]{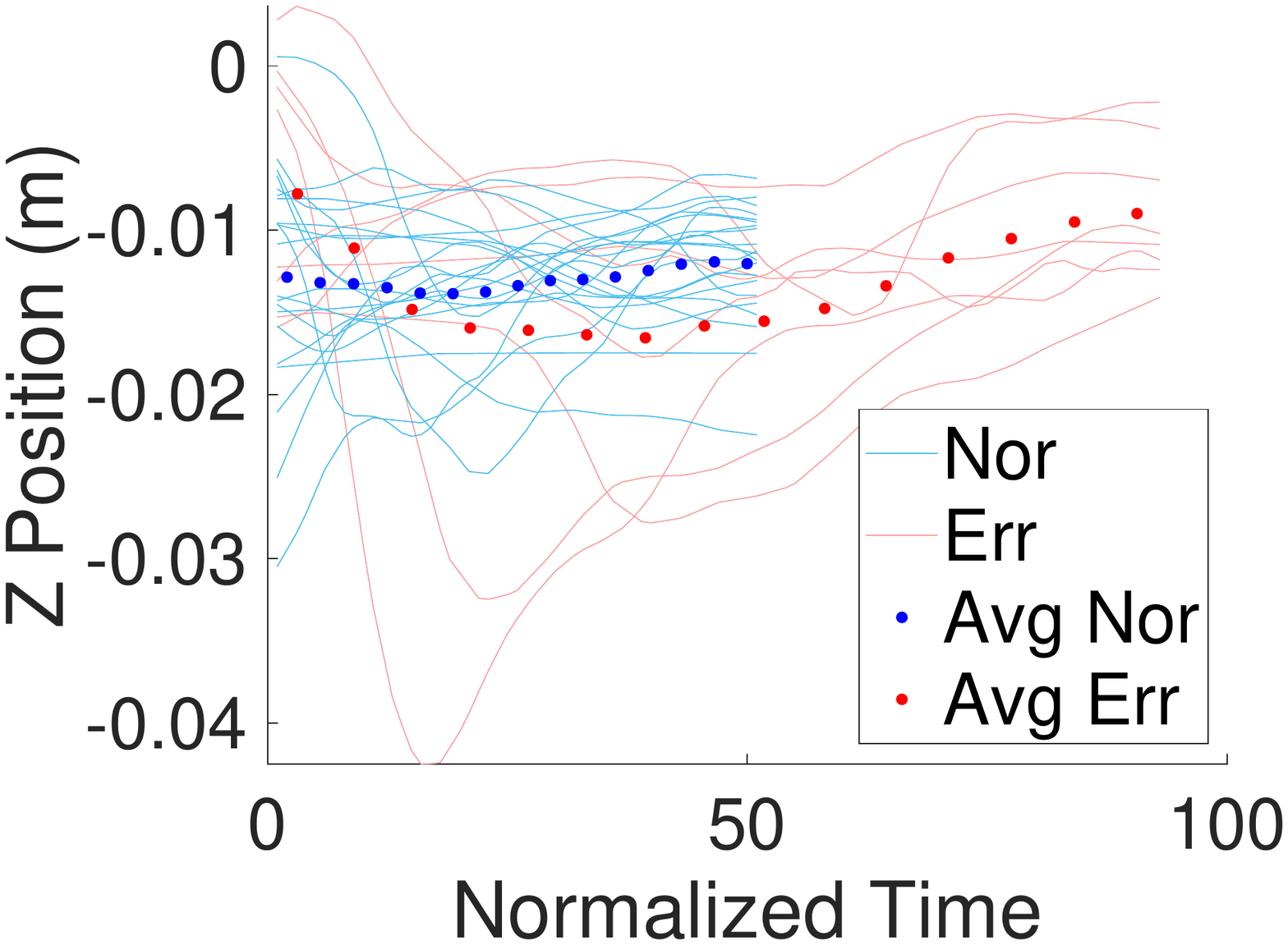}
    \caption{Right Tooltip XYZ Position for Normal and Erroneous G1 in Suturing}
    \label{fig:S_G1_Rpos}
\end{figure}

\item For G2 in Needle Passing, Figure \ref{fig:S_NP_G2_KLD} shows a large difference in KL Divergence for Left Rotational and Linear Velocities which may be due to the active role the left hand plays in stabilizing the ring unlike in Suturing. This is an important contextual difference between tasks. 

\begin{figure}
    \centering
    \begin{subfigure}[b]{0.49\textwidth}
        \centering
        \includegraphics[trim = 0.3in 1.5in 0.3in 1.5in, width=\textwidth]{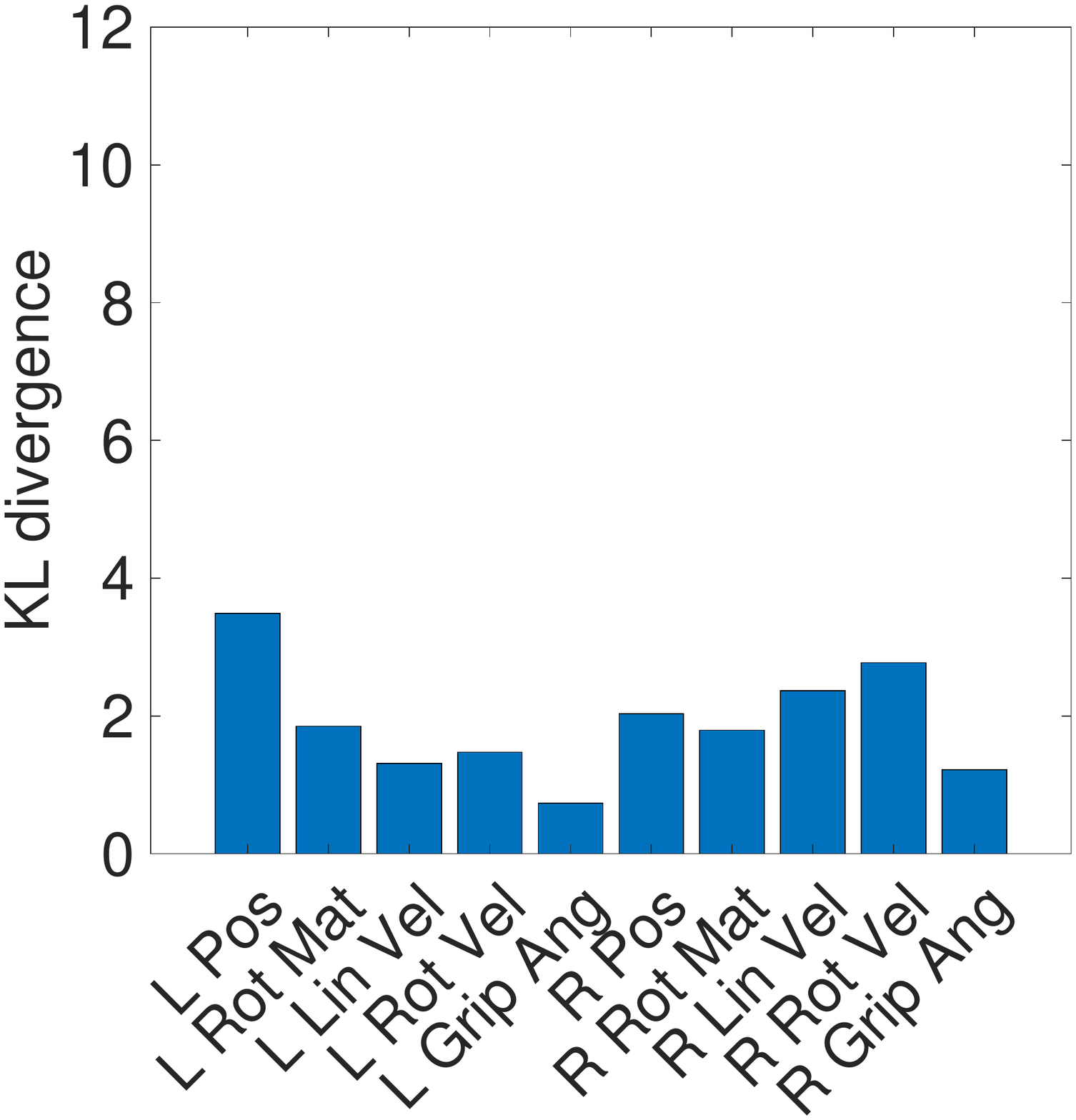}
        \caption{Suturing}
        \label{fig:S_G2_KLD_a}
    \end{subfigure}
    \hfill
    \begin{subfigure}[b]{0.49\textwidth}
        \centering
        \includegraphics[trim = 0.3in 1.5in 0.3in 1.5in, width=\textwidth]{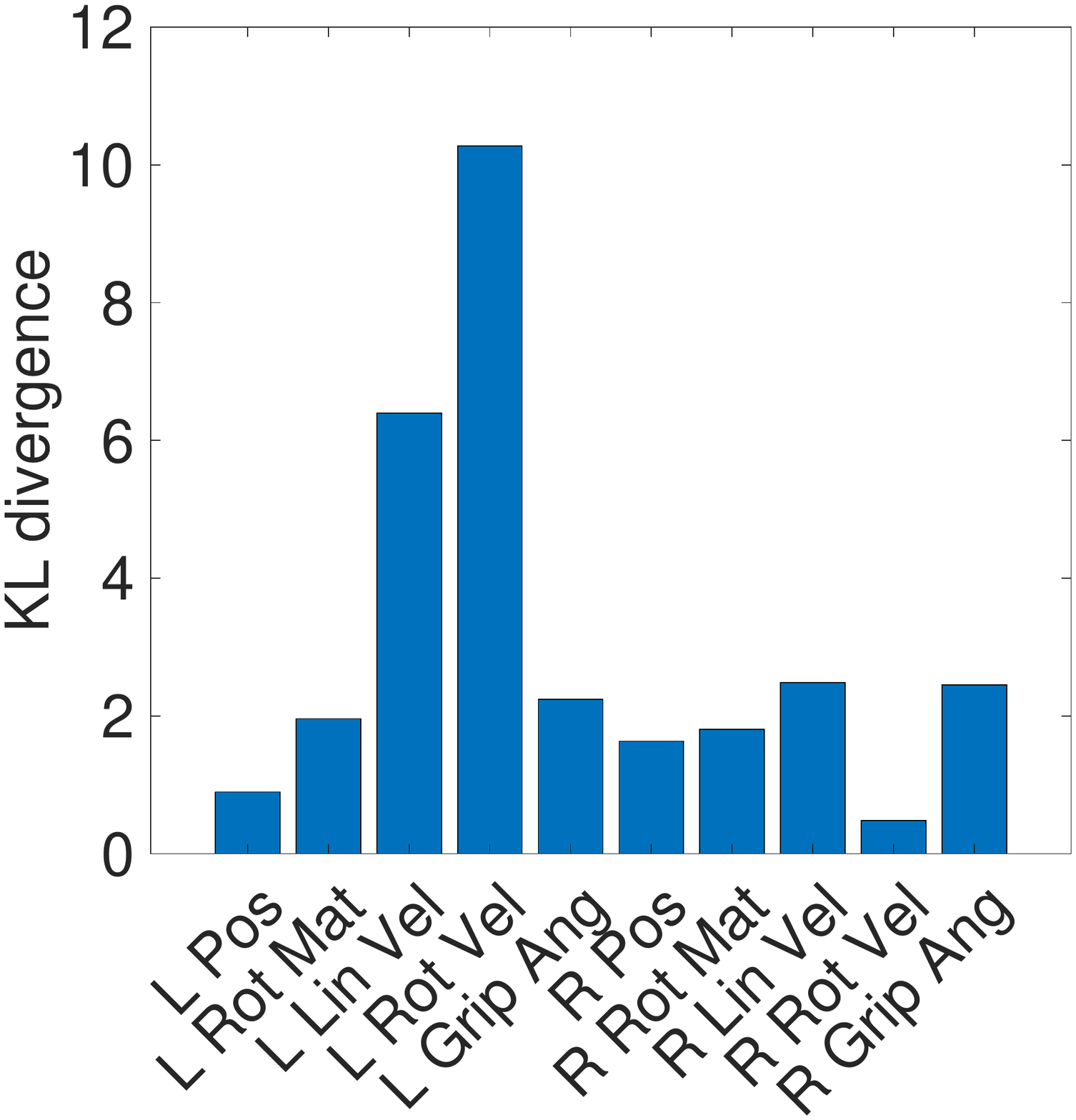}
        \caption{Needle Passing}
        \label{fig:S_G2_KLD_b}
    \end{subfigure}
    \caption{KL Divergence of Kinematic Parameters for G2}
    \label{fig:S_NP_G2_KLD}
\end{figure}

\item The main error mode for G3 was ``Not moving along the curve/ Multiple attempts".
Erroneous gestures in Suturing were caused by lateral, instead of characteristically rotational, movements of the needle while in the fabric. In surgery, lateral movements may tear tissue and contribute to a safety-critical event. This explains the high KL Divergences for the parameters listed in Table \ref{tab:kin_params} and shown in Figure \ref{fig:S_G3_KLD} and is consistent with \cite{sharon2021rate} who found that the rate of orientation change during needle insertion (i.e. Rotational Velocity during G3) was higher for experienced surgeons.

However, Needle Passing shows nearly the opposite result in Figure \ref{fig:NP_G3_KLD}. 
Upon reviewing the gesture clips for both tasks, we noticed that clips for Suturing showed the right grasper driving the needle through the fabric and the left grasper pulling it through, but clips for Needle Passing began with the needle halfway through the ring and only showed the left grasper pulling the needle through. Due to the large difference in KL Divergences between the two tasks, we see that the part of G3 that involves driving the needle with the right grasper is important to this gesture's correct execution.

\begin{figure}
    \centering
    \begin{subfigure}[b]{0.49\textwidth}
        \centering
        \includegraphics[trim = 0.3in 1.5in 0.3in 1.5in, width=\textwidth]{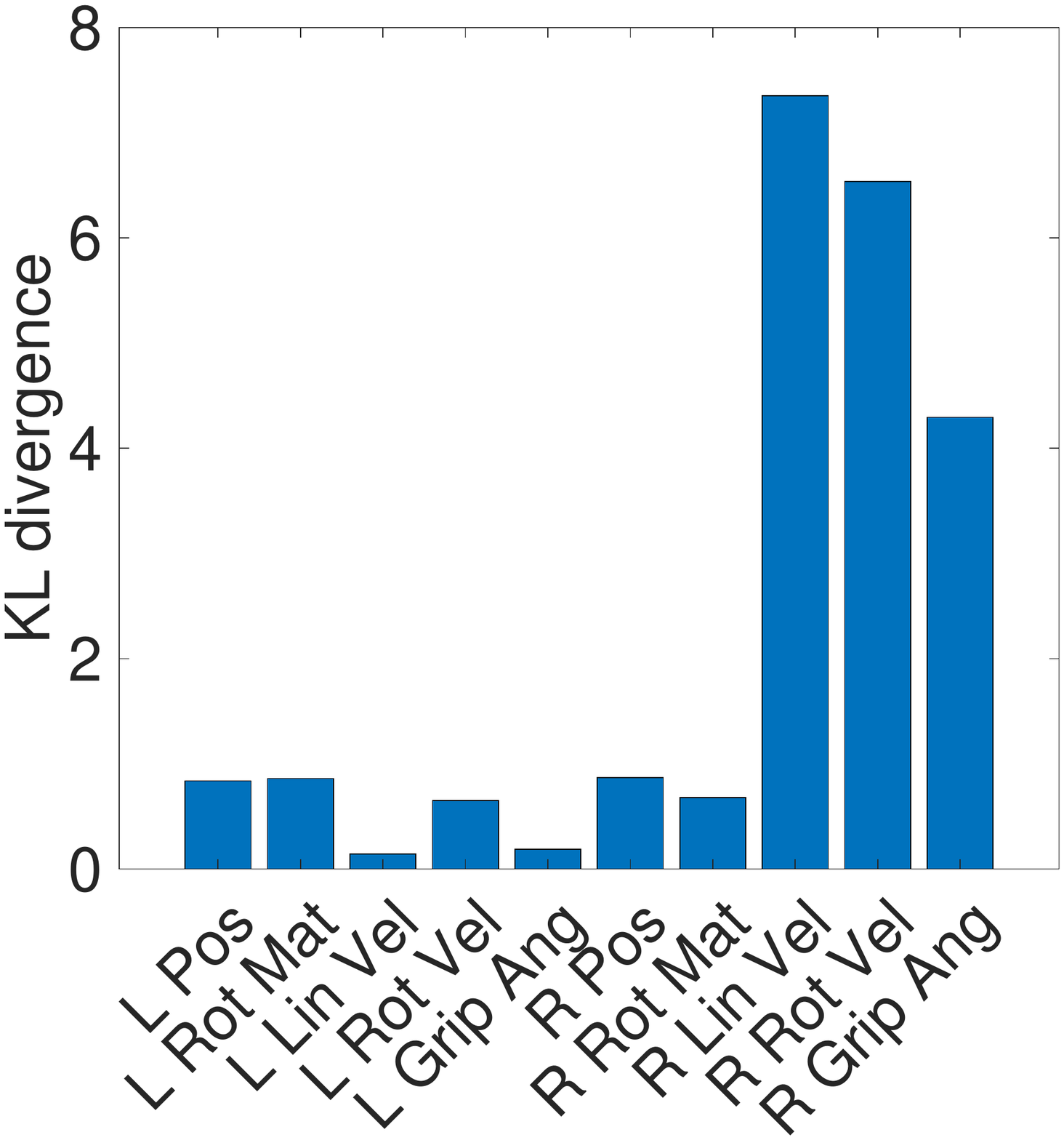}
        \caption{Suturing}
        \label{fig:S_G3_KLD}
    \end{subfigure}
    \hfill
    \begin{subfigure}[b]{0.49\textwidth}
        \centering
        \includegraphics[trim = 0.3in 1.5in 0.3in 1.5in, width=\textwidth]{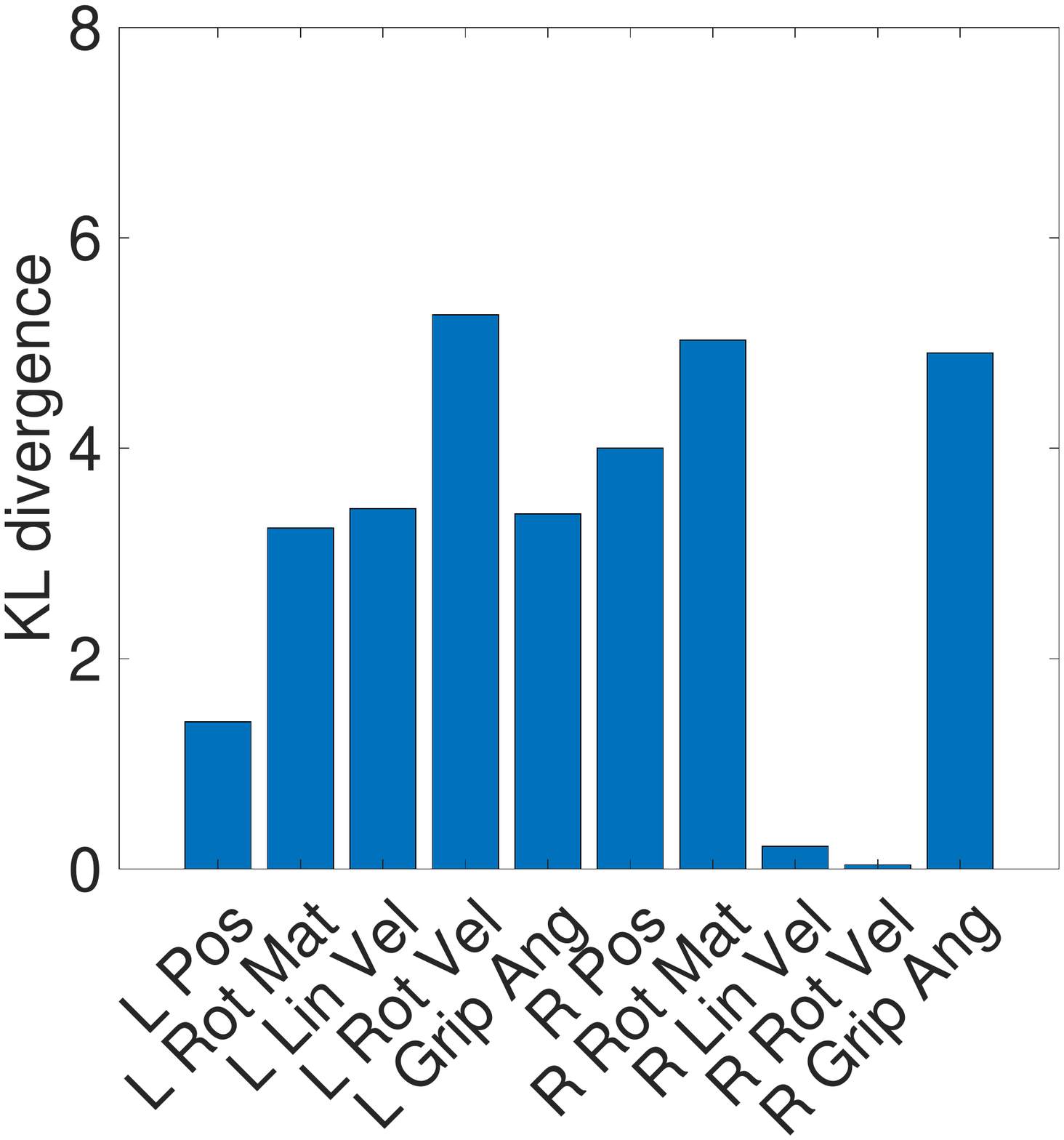}
        \caption{Needle Passing}
        \label{fig:NP_G3_KLD}
    \end{subfigure}
    \caption{KL Divergence of Kinematic Parameters for G3}
    \label{fig:S_NP_G3_KLD}
\end{figure}

\item In both tasks, G4 had KL Divergences below 0.6 for all parameters meaning normal and erroneous examples have very similar kinematics.

\item G6 in Suturing had the most errors with primarily ``Out of view" errors. Figure \ref{fig:S_G6_Lpos} shows that final Y and Z positions for the left grasper were much larger for erroneous gestures as the left grasper exceeded the threshold for visibility while pulling the suture. This explains the large KL Divergence for the Left Position parameter in Figure \ref{fig:S_G6_KLD}.

\begin{figure}
    \centering
    \includegraphics[trim = 0.3in 1.5in 0.3in 1.5in, width=0.45\textwidth]{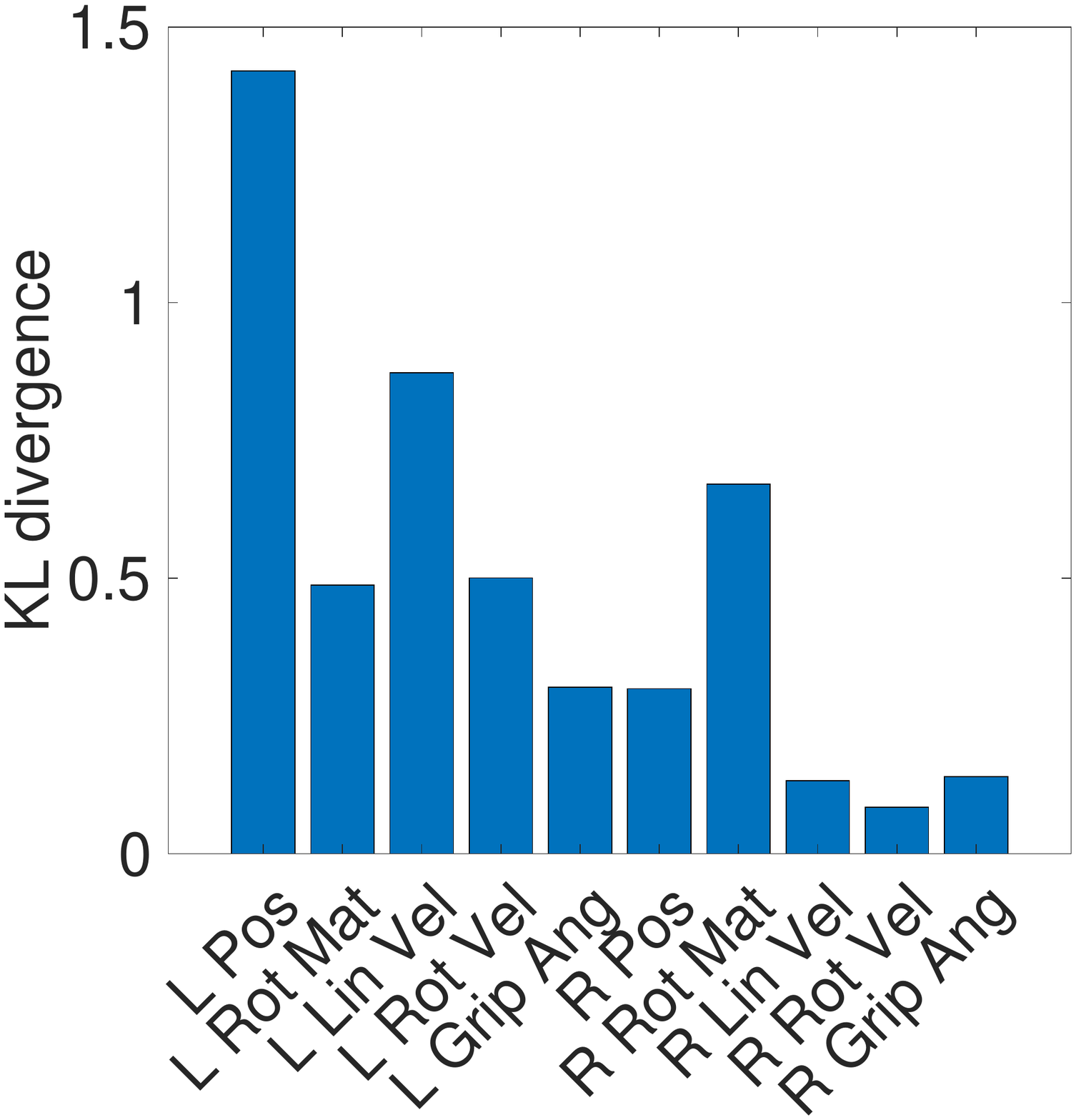}
    \caption{KL Divergence of Kinematic Parameters for G6 in Suturing}
    \label{fig:S_G6_KLD}
\end{figure}

\begin{figure}
    \centering
    \includegraphics[trim = 0.1in 2.5in 0.1in 2.5in, width=0.32\textwidth]{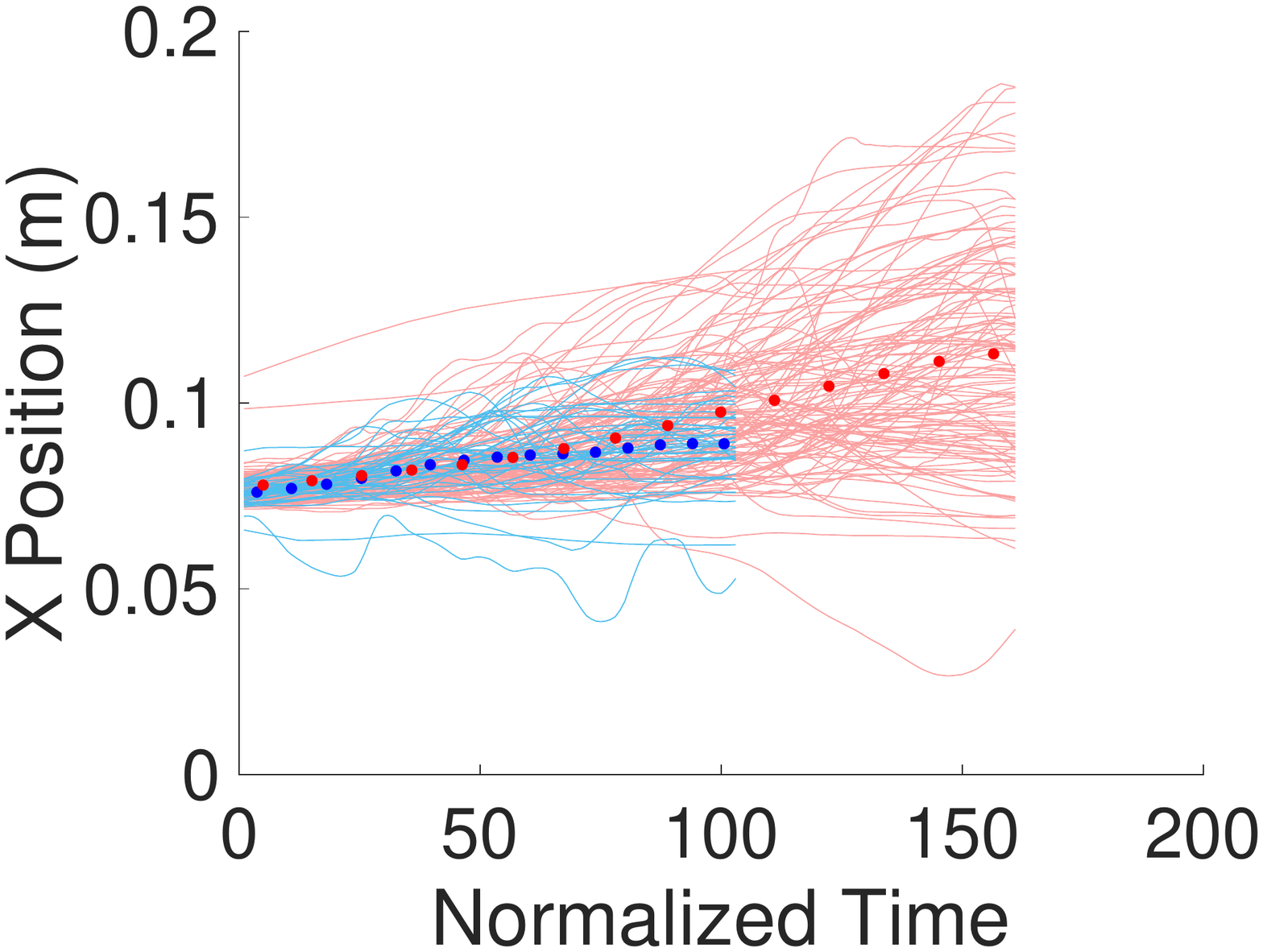}
    \includegraphics[trim = 0.1in 2.5in 0.1in 2.5in, width=0.32\textwidth]{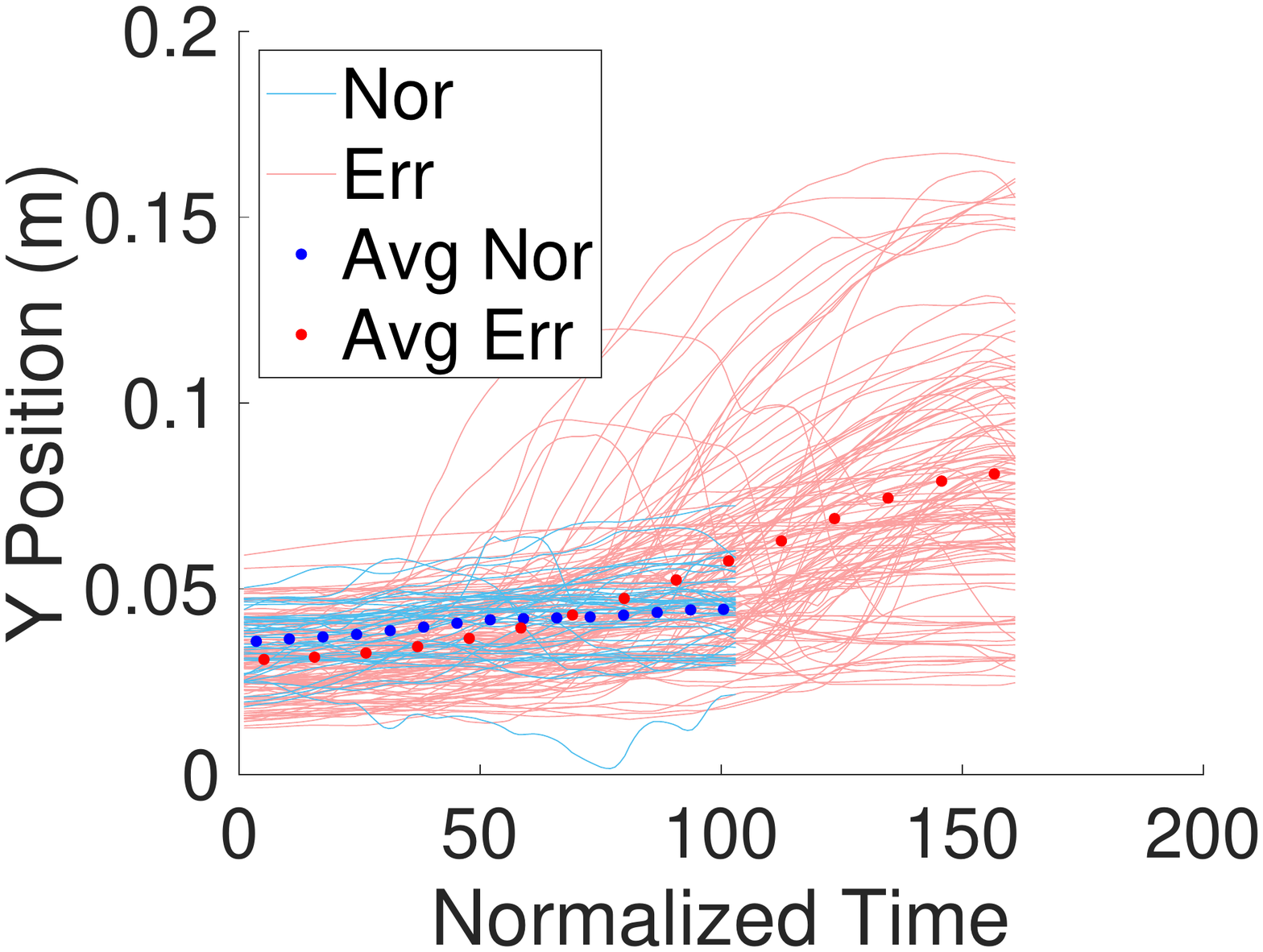}
    \includegraphics[trim = 0.1in 2.5in 0.1in 2.5in, width=0.32\textwidth]{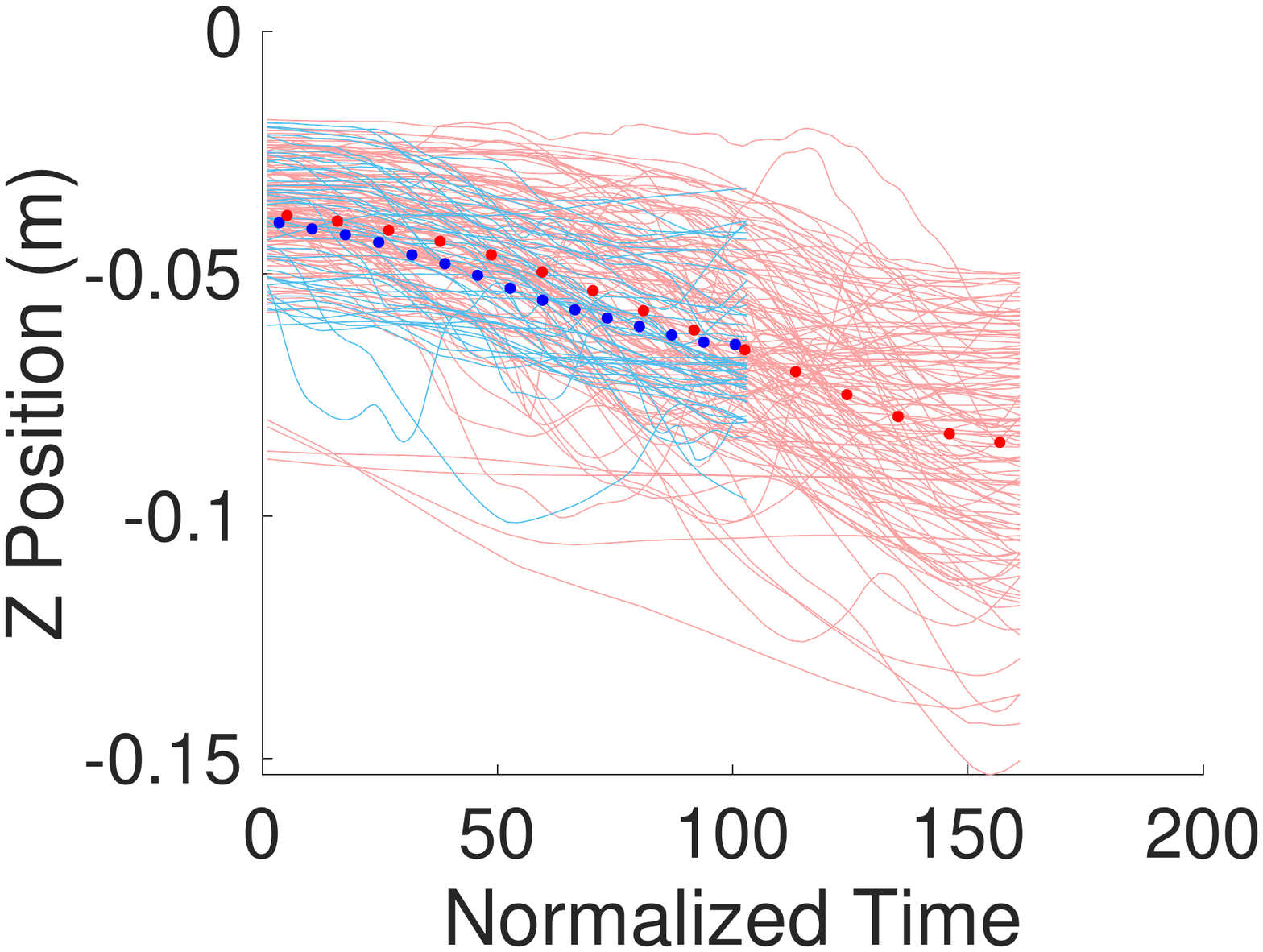}
    \caption{Left Tooltip XYZ Position for Normal and Erroneous G6 in Suturing}
    \label{fig:S_G6_Lpos}
\end{figure}

\item There were two main error modes for G8 in Suturing: ``Multiple attempts" and ``Needle orientation". Figure \ref{fig:S_G8_KLD} shows a comparison of DTW and KL Divergence analysis for G8 from Suturing for all errors, for ``Multiple attempts" versus all other examples, and for ``Needle orientation" versus all other examples. The ``Needle orientation" error alone had the greatest KL Divergence and contributed the most to the results for all errors. For the ``Multiple attempts" error, both the Left and Right Position parameters had the highest KL Divergence which suggests that hand coordination is important in this gesture. Since this gesture includes the right gripper moving to grasp the needle, we see that Right Position is an important parameter in the ``Multiple attempts" error both in G1 and G8.
\end{itemize}

\begin{figure}
    \centering
    \begin{subfigure}[b]{0.32\textwidth}
        \centering
        \includegraphics[trim = 0.3in 1.5in 0.3in 1.5in, width=\textwidth]{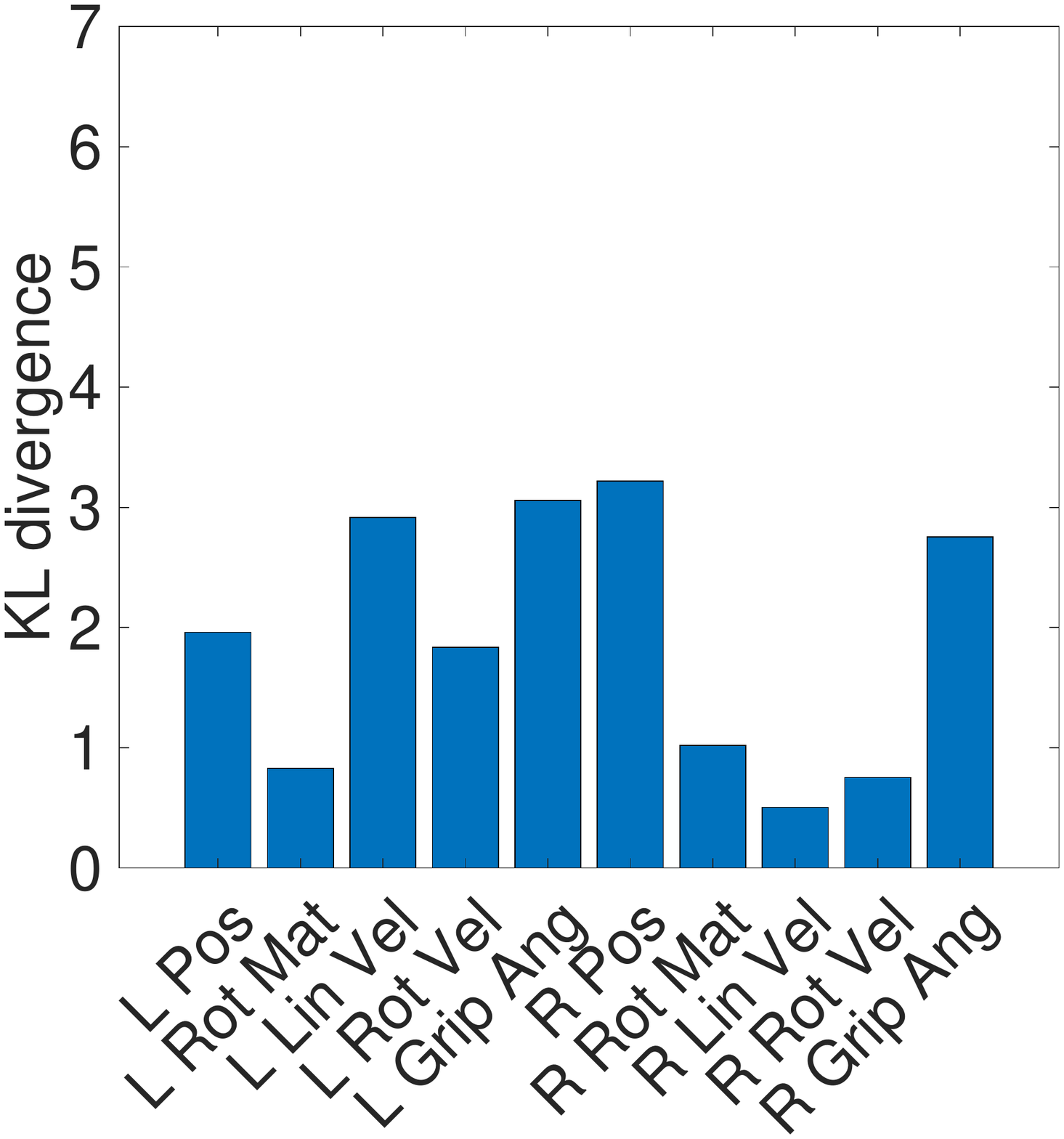}
        \caption{All Errors}
        \label{fig:S_G8_KLD_a}
    \end{subfigure}
    \hfill
    \begin{subfigure}[b]{0.32\textwidth}
        \centering
        \includegraphics[trim = 0.3in 1.5in 0.3in 1.5in, width=\textwidth]{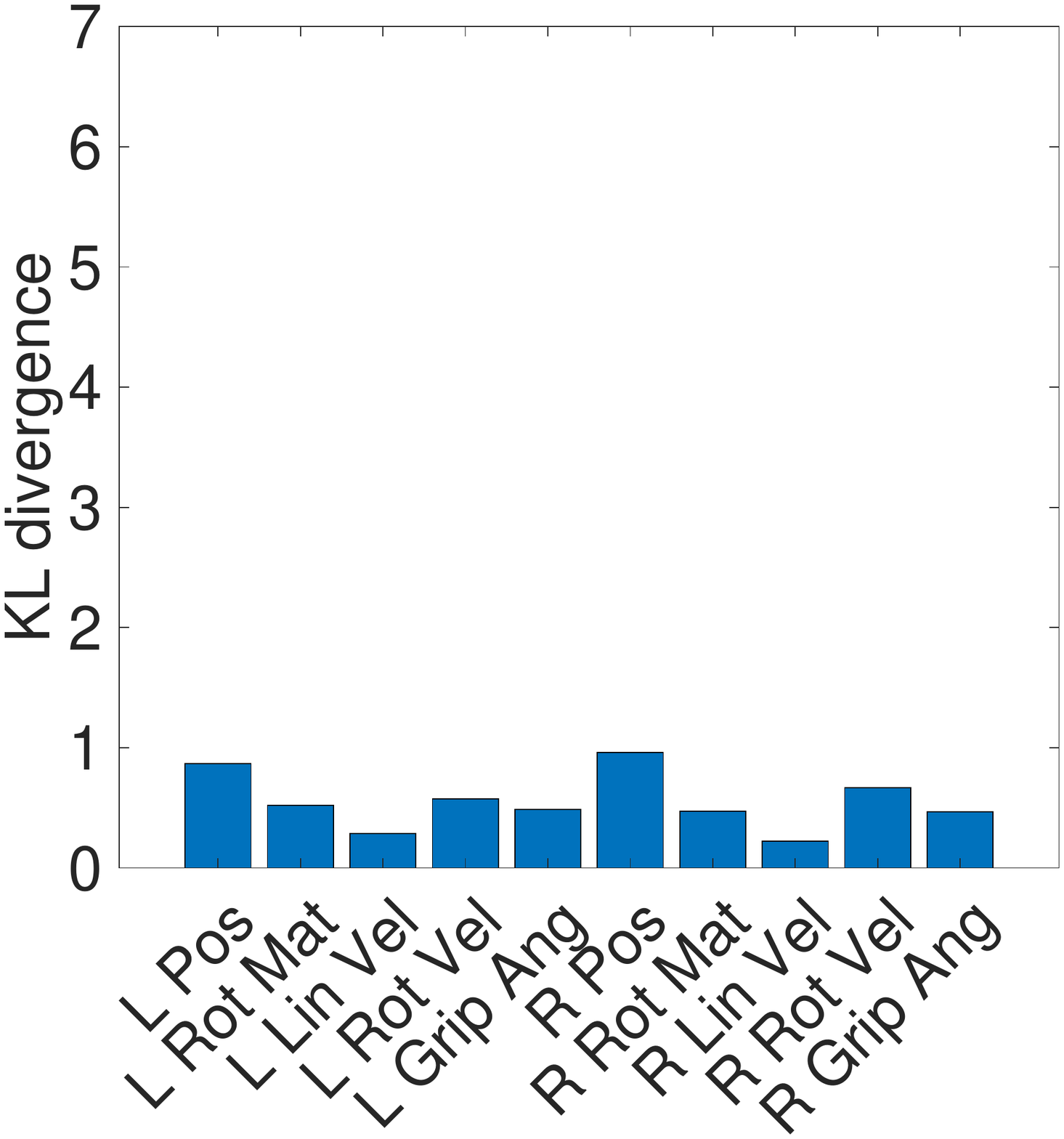}
        \caption{Multiple Attempts Error}
        \label{fig:S_G8_attempts_KLD_b}
    \end{subfigure}
    \hfill
    \begin{subfigure}[b]{0.32\textwidth}
        \centering
         \includegraphics[trim = 0.3in 1.5in 0.3in 1.5in, width=\textwidth]{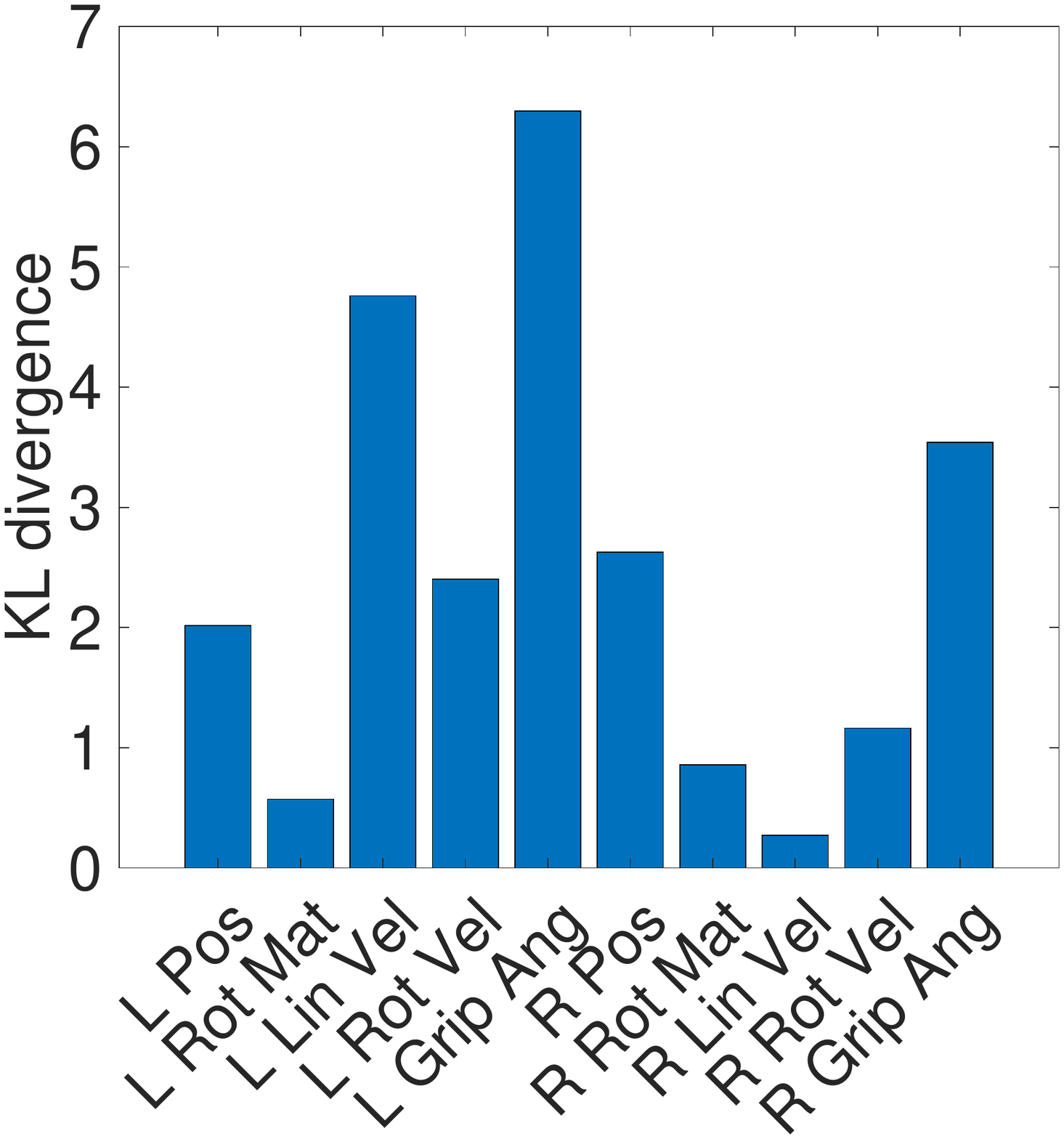}
        \caption{Needle Orientation Error}
        \label{fig:S_G8_attempts_KLD_c}
    \end{subfigure}
    \caption{KL Divergence of Kinematic Parameters for G8 in Suturing} 
    \label{fig:S_G8_KLD}
\end{figure}

\subsubsection{Executional Errors and Skill Levels}
\begin{figure}
    \centering
    \begin{subfigure}{0.45\textwidth}
    \vspace{2em}
    \includegraphics[trim = 0.1in 2.5in 0.1in 2.5in,width=\textwidth]{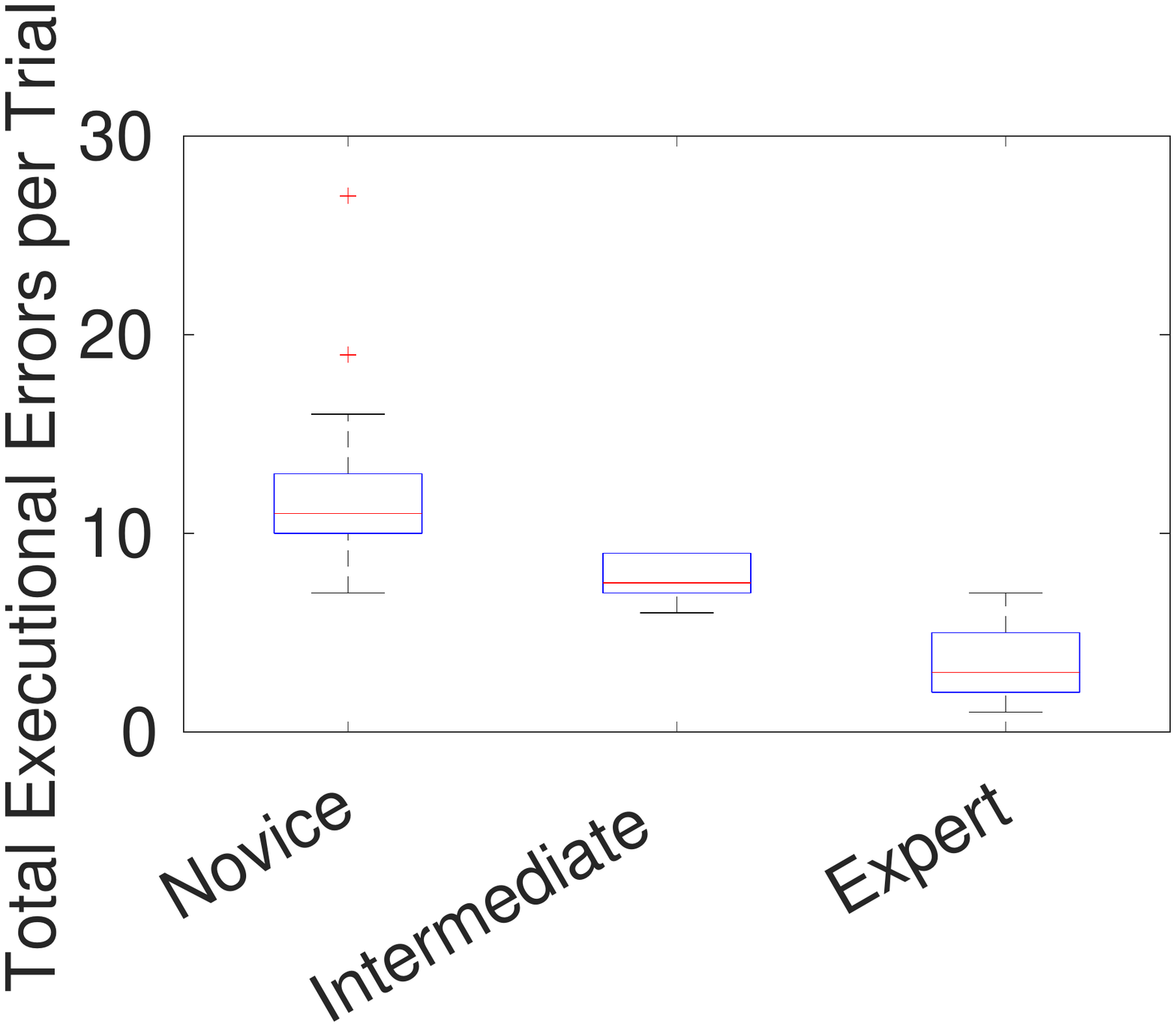}
    \caption{Self-Proclaimed Skill Levels}
    \label{fig:GRS_needlepassing_self_suturing_a}
    \end{subfigure}
    \begin{subfigure}{0.45\textwidth}
    \includegraphics[trim = 0.1in 2.5in 0.1in 1.75in,width=\textwidth]{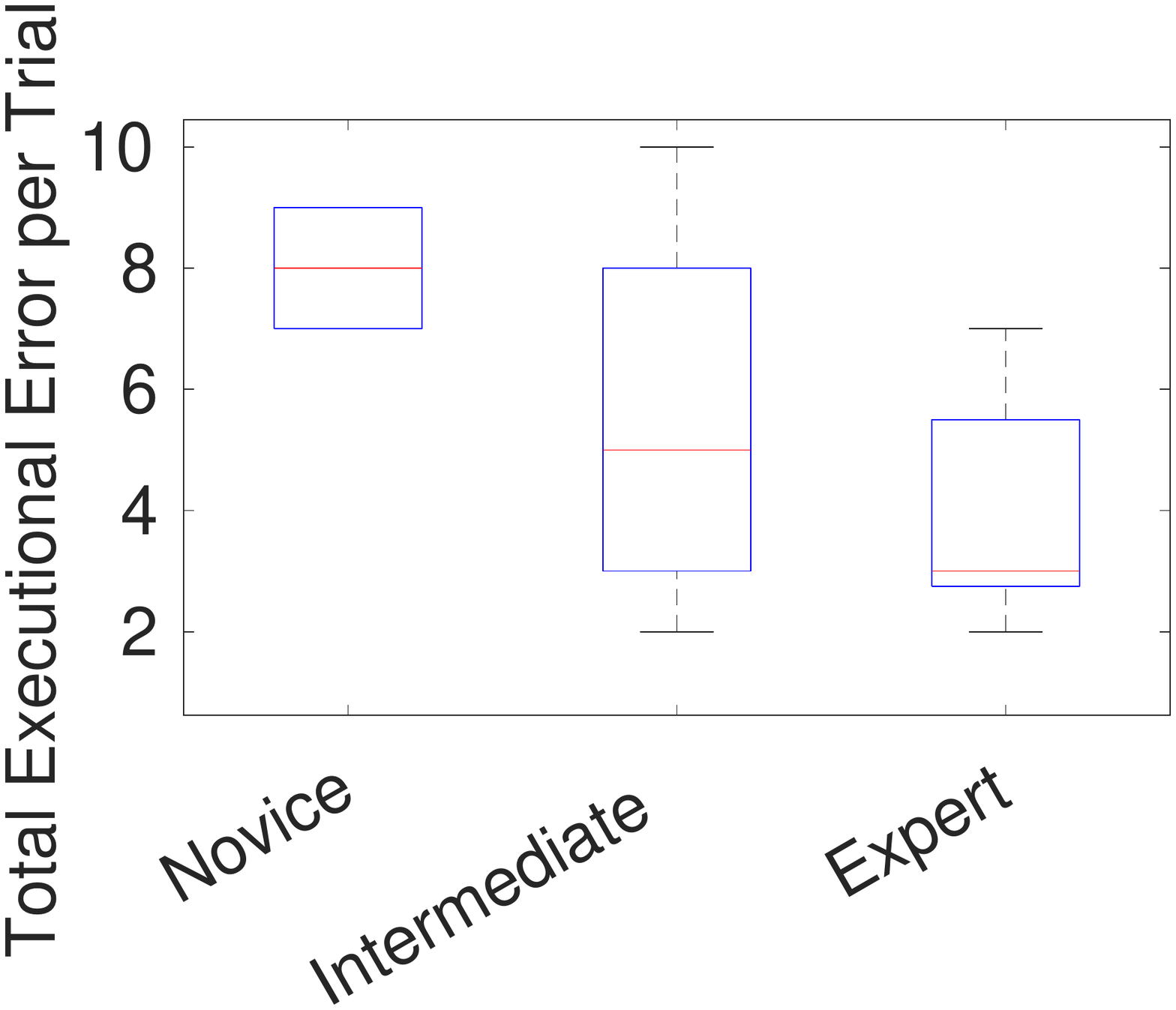}
    \caption{GRS Skill Levels}
    \label{fig:GRS_needlepassing_self_suturing_b}
    \end{subfigure}

    \caption{Total Number of Executional Errors across Surgical Skill Levels: (a) Self-proclaimed Skill Levels in Suturing, (b) GRS Skill Levels in Needle Passing}
    \label{fig:GRS_needlepassing_self_suturing}
\end{figure}

\noindent We analyzed the relationship between executional errors and surgical skill levels. Based on self-proclaimed expertise levels, Figure~\ref{fig:GRS_needlepassing_self_suturing_a} shows a clear difference in the number of errors across different self-proclaimed expertise groups for Suturing. However, no similar pattern was seen in Needle Passing. This might be because Suturing is a more difficult task so the number of executional errors is more reflective of self-proclaimed skill levels in Suturing. 
For GRS-defined skill levels, the total number of executional errors per trial was larger for GRS-Novices than for GRS-Experts in Needle Passing (Figure~\ref{fig:GRS_needlepassing_self_suturing_b}), which is consistent with our expectation that experts with high GRS scores make fewer executional errors than novices. However, since there was only one GRS-Novice trial for Suturing, we did not observe clear differences. 

\subsubsection{Executional Errors and Gesture Duration}
\label{sec:exec_errors_durations}
\begin{figure}
    \centering
    \begin{subfigure}{0.45\textwidth}
    \includegraphics[trim = 0.1in 2.5in 0.1in 2.5in,width=\textwidth]{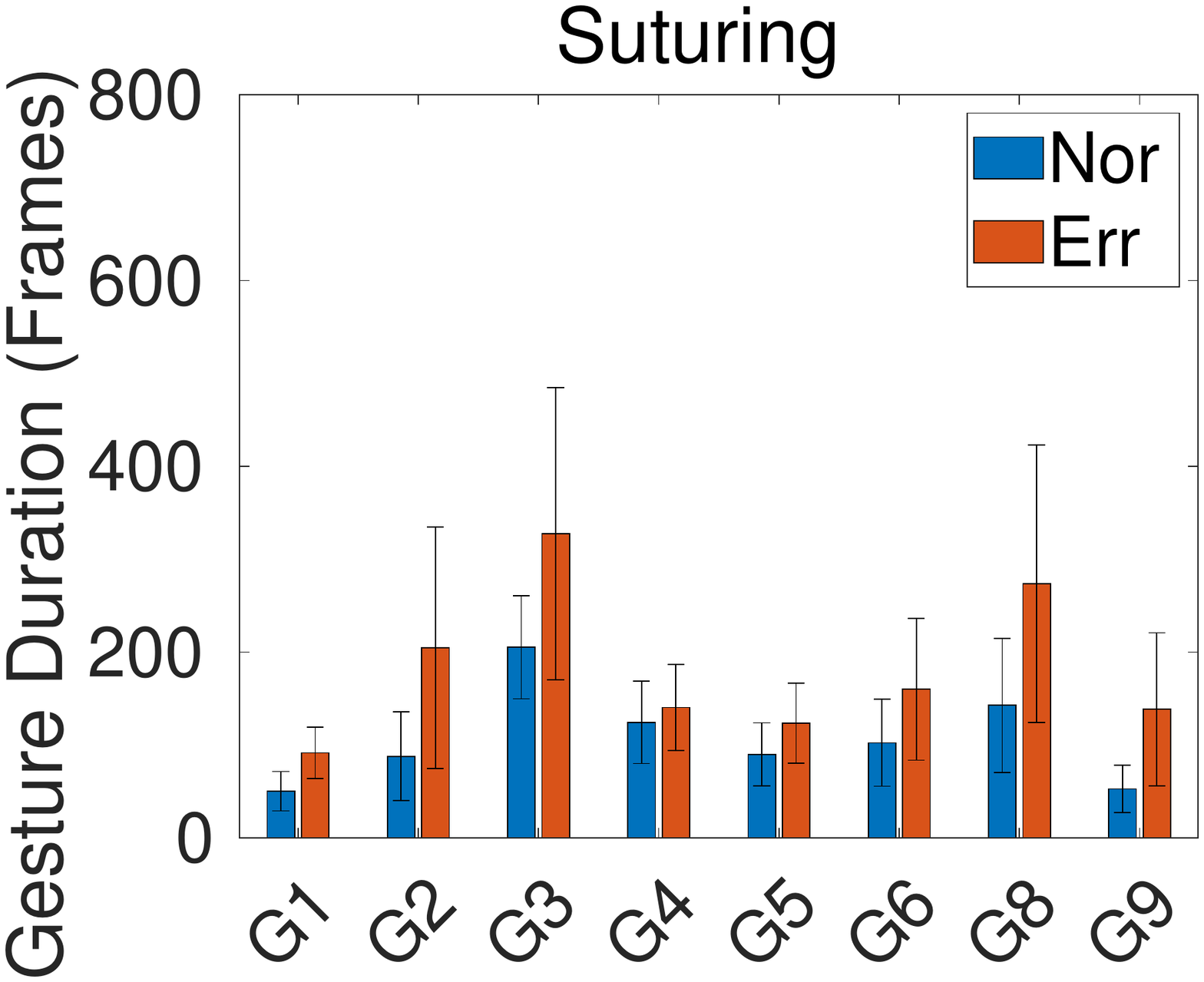}
    \caption{Suturing}
    \end{subfigure}\hfill
    \begin{subfigure}{0.45\textwidth}
    \centering
    \includegraphics[trim = 0.1in 2.5in 0.1in 2.5in,width=\textwidth]{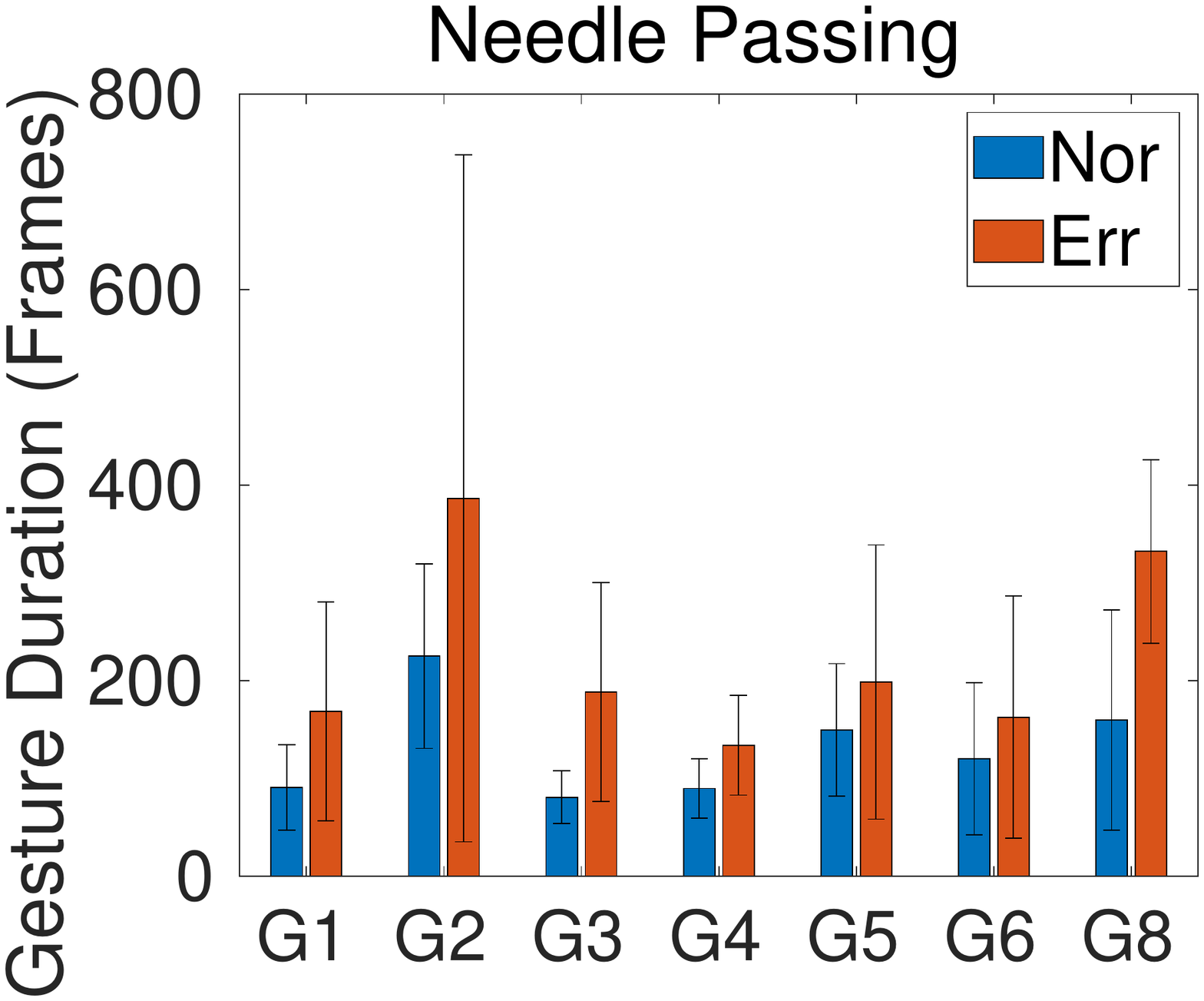}
    \caption{Needle Passing}
    \end{subfigure}\hfill
    \vspace{-1em}
    \caption{Average Normal and Erroneous Gesture Durations}
    \label{fig:G len}
\end{figure}

\begin{figure}[h!]
\centering
\begin{tabular}{cccc}
\includegraphics[trim = 0.1in 2.5in 0.1in 2.5in,width=0.3\textwidth]{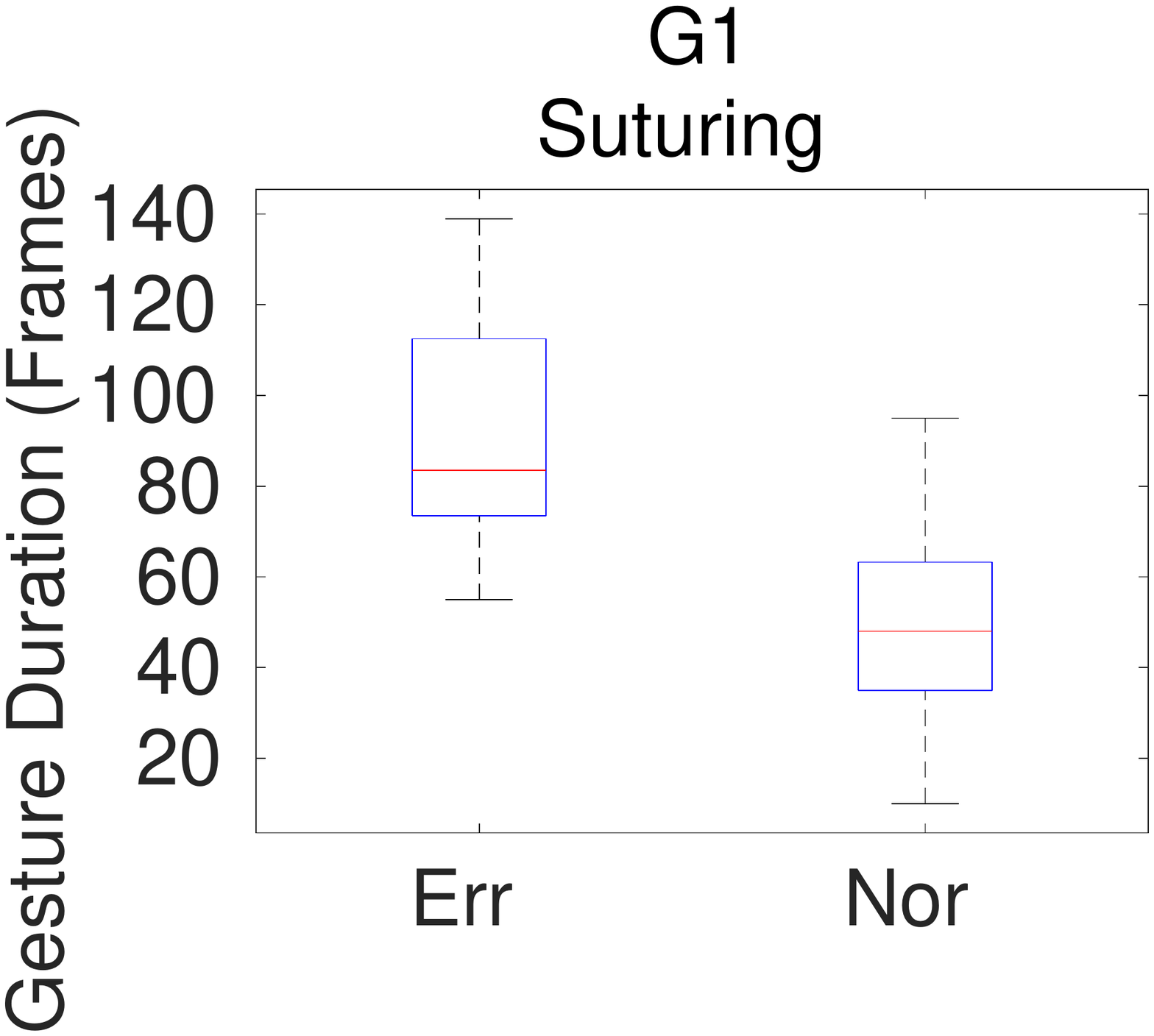} &
\includegraphics[trim = 0.1in 2.5in 0.1in 2.5in,width=0.3\textwidth]{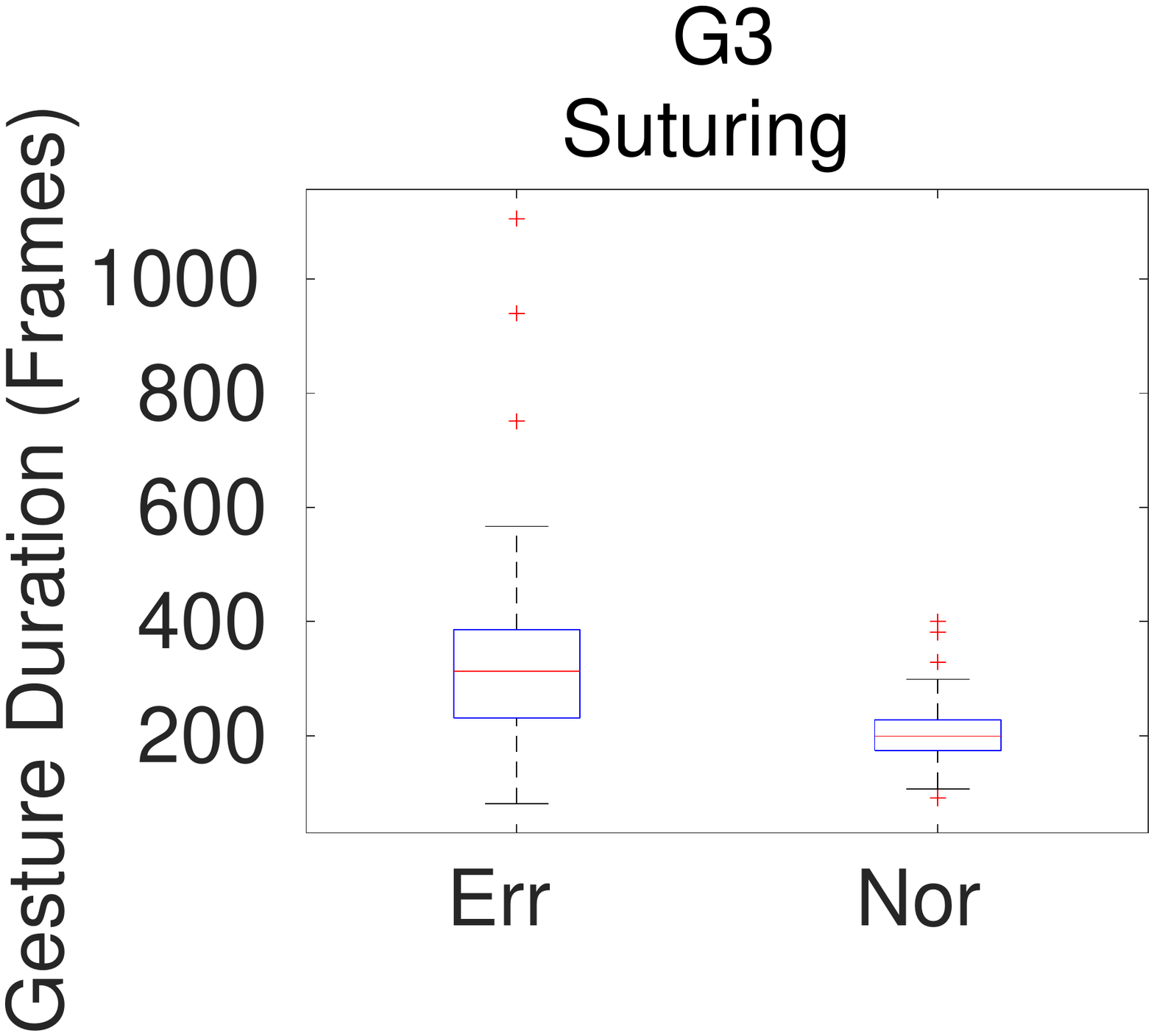} &
\includegraphics[trim = 0.1in 2.5in 0.1in 2.5in,width=0.3\textwidth]{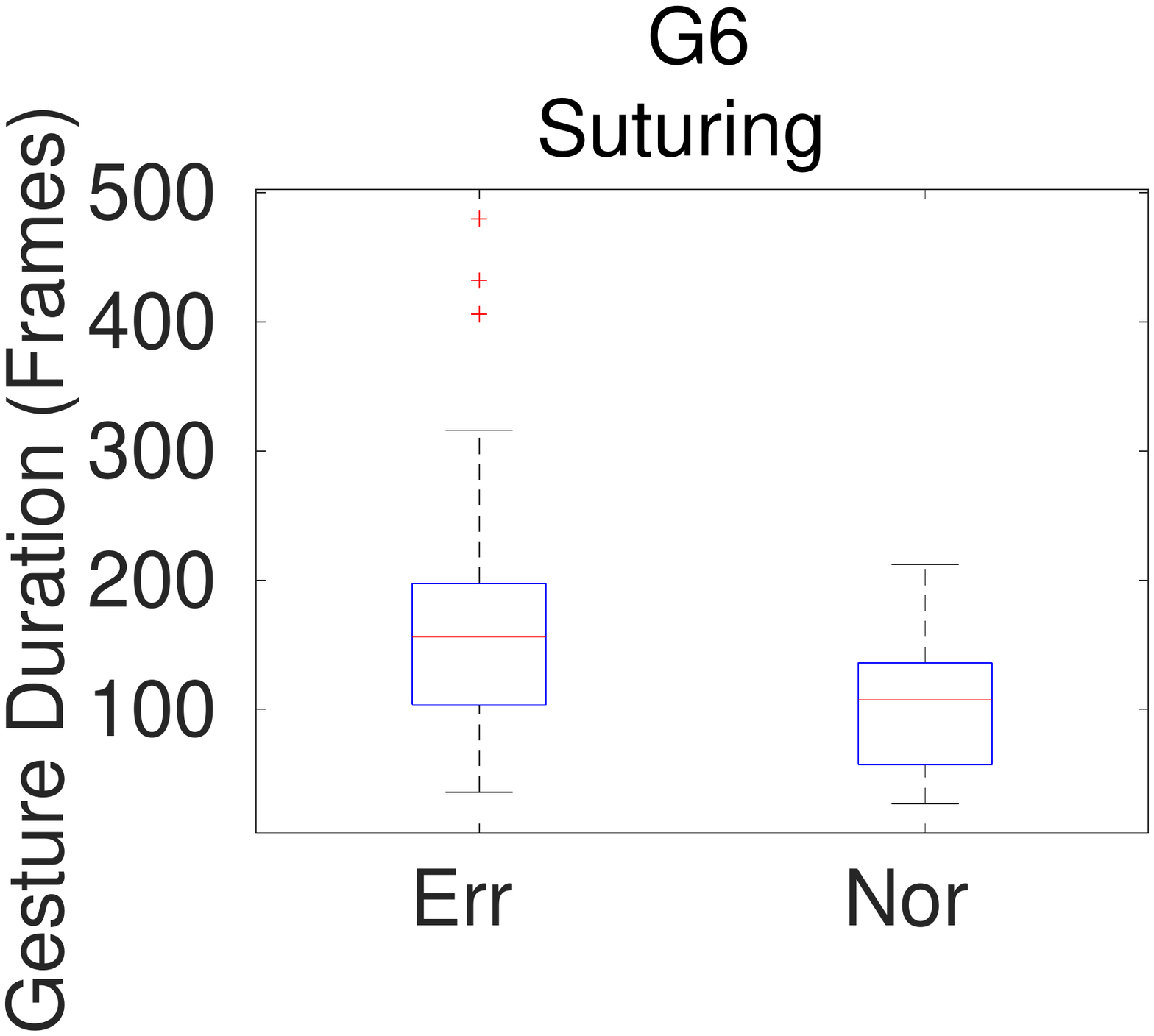} \\
p=9.71e-05  & p=2.36e-10 & p=4.38e-06  \\[6pt]
\end{tabular}
\begin{tabular}{cccc}
\includegraphics[trim = 0.1in 2.5in 0.1in 2.5in,width=0.3\textwidth]{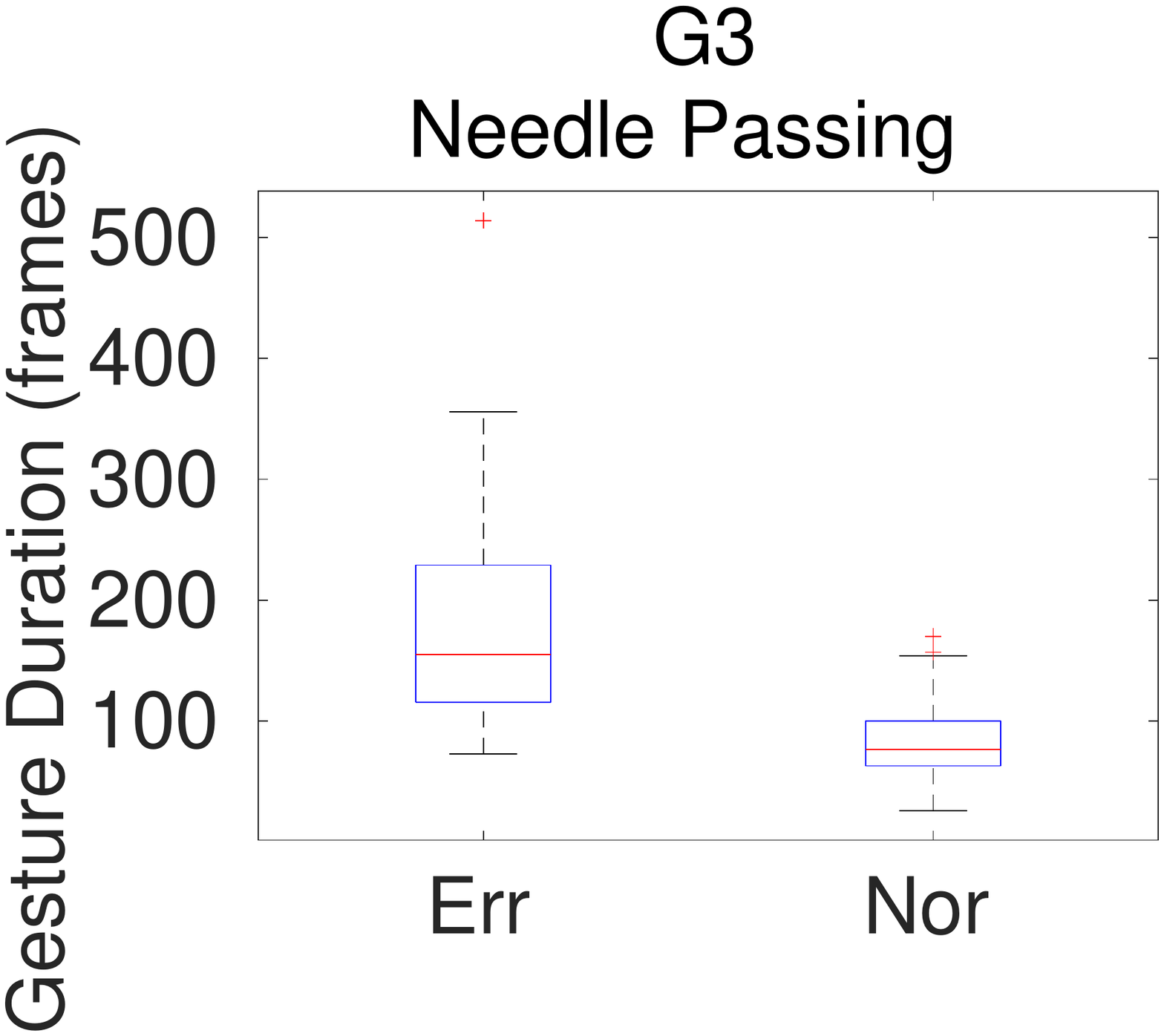} &
\includegraphics[trim = 0.1in 2.5in 0.1in 2.5in,width=0.3\textwidth]{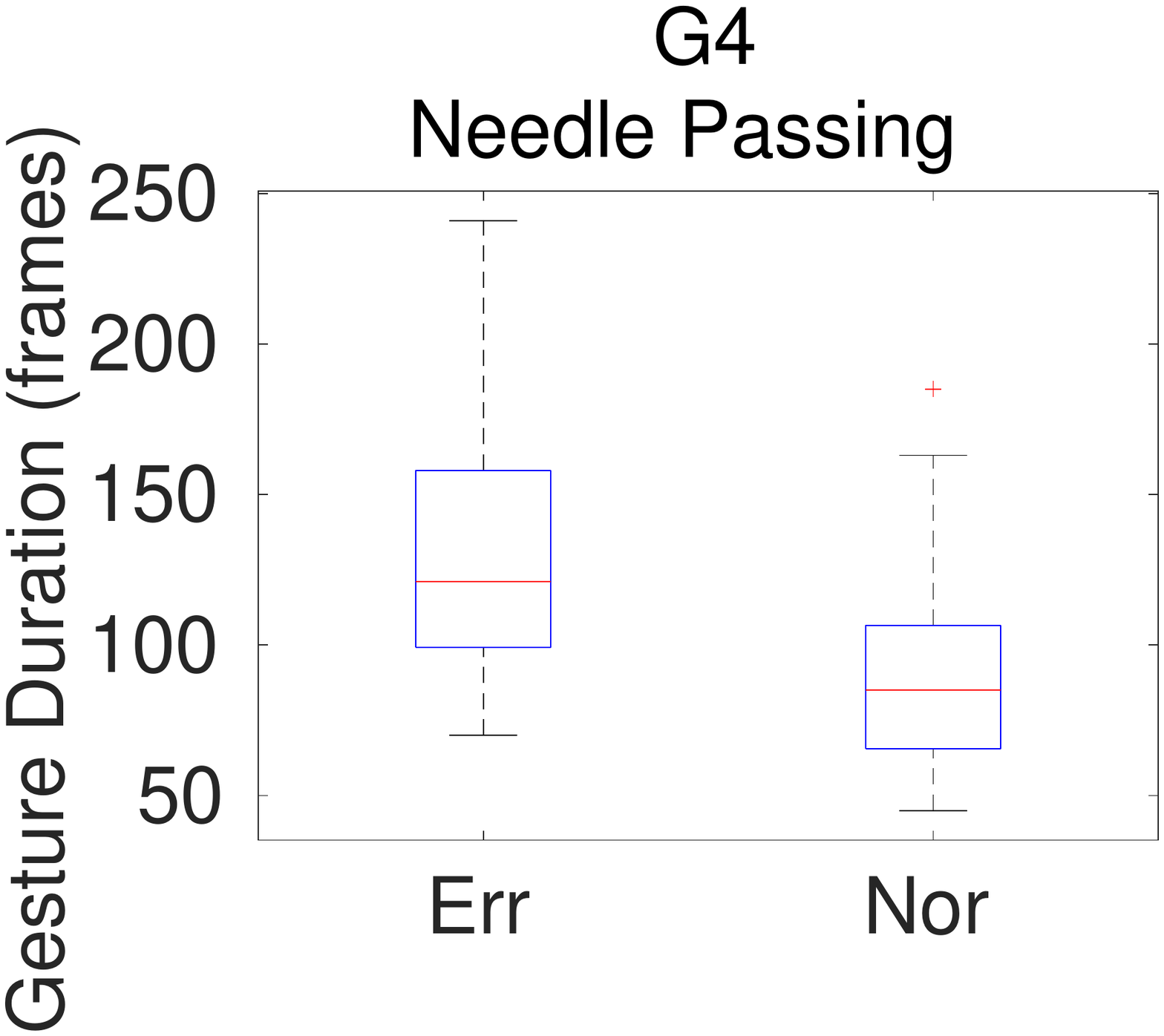} \\
p=2.25e-13  & p=2.96e-06  \\[6pt]
\end{tabular}
\vspace{-1em}
\caption{Erroneous vs. Normal Gesture Durations for Suturing and Needle Passing}
\label{fig:Erroneous vs. Normal Gesture Lengths}
\end{figure}

\noindent We compared erroneous and normal gesture durations using a one-tailed t-test. The null hypothesis is that normal and erroneous gestures have similar durations. The alternative hypothesis is that erroneous gestures are longer than normal gestures. 
Figures \ref{fig:G len} and \ref{fig:Erroneous vs. Normal Gesture Lengths} respectively show average durations and several examples of differences in durations (along with the p-values from the hypothesis test) for normal and erroneous gestures in both tasks.  
\par
 
 We observed that some error types increase the gesture duration, e.g., ``Multiple attempts" for G1, G2, G3, and G8 in Suturing, and G2 and G3 in Needle Passing; and ``Out of view" for G6 and G9 for Suturing, and G4 in Needle Passing. Erroneous gestures with ``Out of view" errors are longer because the distance traveled by the tool is larger, while the speed is similar. 
 We rejected the null hypothesis and found that erroneous gestures are longer than normal gestures for all gestures of both tasks.  
 There is a relatively large p-value (p=0.308) for G4 compared to other p-values.
 This could be because ``Needle orientation" is the primary error mode in G4 and an erroneous needle orientation takes comparable time as a normal needle orientation.

\subsubsection{Executional Errors and Trial Duration}
\begin{figure}[h!]
    \vspace{-0.5em}
    \centering
    \includegraphics[trim = 0.1in 2.5in 0.1in 2.5in,width=0.5\textwidth]{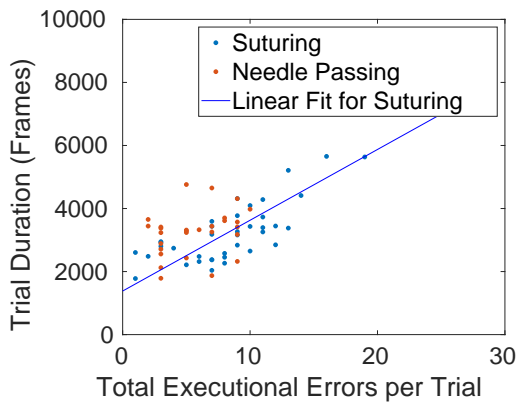}
    \caption{Correlation between Executional Errors and Durations of Trials}
    \label{fig:cor_ge_len_err}
\end{figure}

\noindent For each trial, we summed the executional errors of all gestures in the trial. Then we analyzed the correlation between the total number of executional errors per trial and the duration of the trial (in number of frames). Figure \ref{fig:cor_ge_len_err} shows that there is a significant positive correlation for Suturing (r=0.837, p=6.18e-12), but no significant correlation for Needle Passing. 
This is likely due to the limited number of trials and fewer errors for Needle Passing in the JIGSAWS dataset (see Table \ref{tab:all_errors}).

\textbf{\subsection{Procedural Errors}
\label{sec:procedural_errors_results}}
\noindent We analyzed the numbers and patterns of procedural errors by task, skill level, and subject. 
We hypothesize that the number of procedural errors is inversely proportional to surgical experience and negatively correlated with the demonstration duration. 
\\

\subsubsection{Procedural Errors and Self-Proclaimed Skill Levels}

\begin{table}[t!]
\resizebox{\textwidth}{!}{%
\begin{tabular}{|c|c|c|c|}
\hline
Task - Skill Level         &  \begin{tabular}[c]{@{}l@{}} Total Number of \\Procedural Errors  \end{tabular} &  \begin{tabular}[c]{@{}l@{}} Percentage of \\Erroneous Trials  \end{tabular} & \begin{tabular}[c]{@{}l@{}}Longest Erroneous \\ Gesture Sequences  \end{tabular}                                                    \\ \hline
Suturing - SP-Expert            &11                                                                                 &6/10                                                              & G9-G6-G2 \\ \hline
Suturing - SP-Intermediate      &2                                                                              &2/10                                                                   &  G3-G11             \\ \hline
Suturing - SP-Novice            &23                                                                                &10/19                                                                       & G4-G5-G6-G2        \\ \hline
Needle Passing - SP-Expert      &11                                                                                 & 6/9                                                                       & G2-G6-G10         \\ \hline
Needle Passing - SP-Intermediate &9                                                                                     &5/8                                                                      & G6-G8-G6           \\ \hline
Needle Passing - SP-Novice      &7                                                                              & 4/11                                                                    & G6-G5-G6           \\ \hline

Total &                          63                                                    &33/67                                                                     & -     \\ \hline
\end{tabular}%
}
\caption{Procedural Errors and \textcolor{black}{Self-Proclaimed} Skill Levels}
\label{tab:proce-2task}
\end{table}

\noindent We compared the percentage of erroneous trials for SP-Novice, SP-Intermediate and SP-Expert groups. As shown in Table \ref{tab:proce-2task}, we observe that for both tasks, SP-Expert surgeons on average had more procedural errors compared to SP-Intermediate surgeons. For Needle Passing, SP-Intermediate surgeons made more errors than SP-Novice surgeons. This could be due to variations in surgical style especially in more experienced surgeon groups. For example, our analysis of error patterns by subject showed that one of the SP-Expert subjects consistently made G9-G11 transitions in different trials of Suturing (see Figure \ref{fig:hierarchy}). This is a unique non-safety-critical pattern that was not observed in the trials by other subjects. However, procedural errors by SP-Novice subjects were more random and did not follow specific patterns.   

Of the two tasks, the longest erroneous gesture sequence is G4-G5-G6-G2 in Suturing performed by an SP-Novice surgeon. Upon review of the video, G5 may be a typo in the transcript. \\

\begin{table}[]
\centering
\resizebox{0.85\textwidth}{!}{%
\begin{tabular}{|l|l|l|l|l|}
\hline
 & \multicolumn{2}{l|}{Suturing} & \multicolumn{2}{l|}{Needle Passing} \\ \hline
GRS Sub-score & \begin{tabular}[c]{@{}l@{}}Correlation \\ Coefficient\end{tabular} & p-value & \begin{tabular}[c]{@{}l@{}}Correlation \\ Coefficient\end{tabular} & p-value \\ \hline
Respect for Tissue & -0.41 & 0.009
& -0.12 & 0.528 \\ \hline
\begin{tabular}[c]{@{}l@{}}Suture \& Needle \\ Handling\end{tabular} & -0.50 & 0.001
& -0.26 & 0.184 \\ \hline
Time \& Motion & -0.55 & 
\textless0.001 & -0.11 & 0.594 \\ \hline
Flow of Operation & -0.43 & 0.006
& -0.22 & 0.268 \\ \hline
\begin{tabular}[c]{@{}l@{}}Overall \\ Performance\end{tabular} & -0.62 & 
\textless0.001 & -0.16 & 0.412 \\ \hline
\begin{tabular}[c]{@{}l@{}}Quality of \\ Final Product\end{tabular} & -0.26 & 0.115 & -0.02 & 0.920 \\ \hline
GRS Score & -0.51 & 
\textless0.001 & -0.15 & 0.434 \\ \hline
\end{tabular}%
}
\caption{Correlation between Number of Procedural Errors and GRS sub-scores for Suturing and Needle Passing}
\label{tab:coef_procedural_GRS}
\end{table}


\subsubsection{Procedural Errors and GRS Skill Levels}
\noindent We analyzed the correlation between the number of procedural errors and GRS score (Table \ref{tab:coef_procedural_GRS}).  
The strongest negative correlation between the number of procedural errors, GRS score, and GRS sub-scores is in Suturing.  
Among the sub-scores of Suturing, Overall Performance has the strongest negative correlation with procedural errors. This could happen because an inefficient procedure has the greatest impact on Overall Performance in Suturing. Needle Passing has a weaker negative correlation between procedural errors and GRS score.  
The Needle Handling sub-score has the highest negative correlation with the number of procedural errors. This is expected as Needle Handling is the main component of the Needle Passing task and poor performance due to procedural errors will lead to a lower score.  

\subsubsection{Procedural Errors and Trial Duration}
\begin{table}[t!]
\centering
\begin{tabular}{|l|l|l|l|}
\hline
Task & r & p-value  \\ \hline
Suturing & 0.71 & \textless0.001\\ \hline 
Needle Passing & 0.17 & 0.399  \\ \hline
\end{tabular}
\caption{Correlation between Procedural Errors and Trial Durations}
\label{tab:procedure errors vs length}
\end{table}

\noindent In Suturing, there is a significant positive correlation between procedural errors and the duration of the trials, so more procedural errors lead to longer trials.  
However, there is no significant correlation in Needle Passing 
possibly because Needle Passing is an easier task (Table \ref{tab:procedure errors vs length}). 

\section{Discussion}
\noindent We used our insights from the analysis of executional and procedural errors in the JIGSAWS dataset to answer the research questions posed in Section~\ref{sec:methods}:

\shortsection{RQ1: Which tasks and gestures are most prone to errors?}
More challenging gestures in each task that require a high level of accuracy and hand coordination were more prone to executional errors. As shown in Table \ref{tab:all_errors}, Suturing is more difficult than Needle Passing and had a greater number of executional errors. G6, G3, and G4 had the greatest number of executional errors in Suturing while G2 and G6 had the greatest number of executional errors in Needle Passing. 

However, procedural errors were almost equally likely in both tasks. 18/39 Suturing trials and 15/28 Needle Passing trials contained procedural errors (Table \ref{tab:proce-2task}). 

\shortsection{RQ2: Are there common error modes or patterns across gestures and tasks?}
Within each task, each gesture had a different predominant error mode that correlated with the challenging aspects of performing the gesture. For both tasks, G2 and G3 had a large number of ``Multiple attempts" errors, G5 had the fewest errors, G6 had the largest number of ``Out of view" errors, and G4 and G6 had the greatest number of gestures with Multiple Errors. Thus, executional errors are context-specific and their type and frequency depend on both task and gesture. 

\shortsection{RQ3: Are erroneous gestures distinguishable from normal gestures?}
KL Divergence magnitude provides insight into which gestures have the greatest difference between normal and erroneous examples. We found that G9 from Suturing, G2 from Needle Passing, and G3 from Suturing had the three greatest KL Divergences for any parameter. However, upon examination of kinematic data for the Left Gripper Angle of G9 from Suturing, the large KL Divergence for this gesture could be due to the effect of three outlying gestures on an already relatively small sample of only 24 examples. 

\shortsection{RQ4: What kinematic parameters can be used to distinguish between normal and erroneous gestures?}
Table \ref{tab:kin_params} lists the parameters with the greatest KL Divergence for each gesture and task which \textcolor{black}{
can be used to develop automated error detectors.
Focusing on a subset of variables for a given task and gesture may enable and improve real-time error detection and skill-assessment by reducing processing time and providing context.}
Our KL divergence analysis approximated the DTW distance distributions as Gaussian, which might not be always accurate. Future work will focus on further refining our analysis method to address this limitation.

\shortsection{RQ5: Do errors impact the duration of the trajectory?} 
Executional and procedural errors often lead to lengthier trials, especially during more complicated tasks such as Suturing. Timely detection and correction during training or surgery will enable more efficient and safer patient care, and aid in reducing learning curves and time to certification. 

\shortsection{RQ6: Are there any correlations between errors and surgical skill levels?}
The total number of executional errors made per trial could help differentiate skill levels. We found this to be true for self-proclaimed skill levels in Suturing and GRS skill levels in Needle Passing. 

There was a significant negative correlation between overall GRS scores and sub-scores and the total number of procedural errors made per trial in Suturing meaning a greater number of procedural errors contributes to a lower GRS score. 
\textcolor{black}{After examining procedural error patterns, 
we noticed that self-proclaimed novice surgeons tend to closely follow the grammar graph, but experts have unique signatures that deviate from the graph. This motivates developing automated gesture identification and procedural error detection techniques based on grammar graphs for training novice surgeons in simulation experiments.} Further verification of the correlation between errors and skill levels requires access to larger datasets representing more tasks and surgeons.
Additionally, the grammar graphs cannot completely capture all possible valid gesture sequences and surgeon-specific signatures, 
\textcolor{black}{and} manual labeling may introduce errors in the gesture transcripts (e.g., incorrectly adding or missing some gestures) that might lead to incorrect detection of errors. 

\section{Conclusion}
\noindent We presented a new rubric and method for objective evaluation of RAS procedures with a focus on gesture and task-specific executional and procedural errors. We used the proposed rubric to evaluate dry-lab demonstrations of Suturing and Needle Passing tasks.
Our analysis identified the most common error modes and their correlations with skill levels and demonstration times as well as important error-specific kinematic parameters that distinguish erroneous gestures. This study is a step towards developing methods for automated error detection and providing real-time context-dependent feedback for performance improvement. \textcolor{black}{Future work will extend the error rubric and analytic methods to larger datasets and other surgical tasks.}

\begin{acknowledgements}
\noindent This work was partially supported by an award from the Engineering-in-Medicine center at the University of Virginia and the National Science Foundation NRT program (Grant No. 1829004). We thank Samin Yasar for his contributions to the initial drafts of the article, and Asha Maran, Devin Gardner, Rohan Chandra, and Ian Reyes for their help with labeling the data.
\end{acknowledgements}

%

\section*{Conflict of interest}
\vspace{-1em}
\noindent The authors declare that they have no conflict of interest.

\section*{Ethical approval}
\vspace{-1em}
\noindent This article does not contain any studies involving human participants performed by any of the authors.

\bibliographystyle{spmpsci}      
\bibliography{ijmrcas.bib}   

%
%

\end{document}